\documentclass[sigconf]{acmart}
\usepackage{booktabs} % For formal tables
\usepackage{verbatim}
\usepackage{graphicx}
\usepackage{multirow}
\usepackage{array}
\usepackage{subfigure}
\usepackage{multicol}
\usepackage{color}
\usepackage{enumitem}
\usepackage{graphicx}
\usepackage{epsfig}
\usepackage{epsfig}
\usepackage{natbib}
\usepackage{amsmath}
\usepackage{xspace}
\usepackage{booktabs}
\usepackage{url}
\usepackage{algorithm}
\usepackage{algorithmic}
\usepackage{bm}
\usepackage{balance}

\newcommand{\hide}[1]{} %hide
\newcommand{\vpara}[1]{\noindent\textbf{#1 }}

\newcommand{\beq}[1]{\begin{equation}#1\end{equation}\vspace{-0.01in}}

\newcommand\expt{\mathbb E}

\newcommand\RR{\mathbb{R}}

\newcommand\nattrs{d}
\newcommand\nsamps{n}
\newcommand\ndim{k}

\theoremstyle{definition}
\newtheorem{definition}{Definition}

\newtheorem{theorem}{Theorem}
\newtheorem{lemma}{Lemma}

\title{
Learning Fair Representations via an Adversarial Framework
}

\author{
Rui Feng$^1$, Yang Yang$^{1\dag}$, Yuehan Lyu$^2$, Chenhao Tan$^3$, Yizhou Sun$^{4}$ and Chunping Wang$^{2}$\\
$^1$Zhejiang University \quad $^2$ PPDAI GROUP Inc.	\\ $^3$University of Colorado Boulder \quad $^4$University of California, Los Angeles	\\
$^\dag$Corresponding author: yangya@zju.edu.cn\\
}

\begin{document}
	
	%!TEX root = main.tex

\begin{abstract}

% The aim of this study is to learn the latent representation of individuals achieving fairness by obstructing information about protected attributes (e.g., race and gender), that may cause bias and discrimination in statistical inference and learning.  

% As classification algorithms have become increasingly used in societal ,
Fairness has become a central issue for our research community as classification algorithms are adopted in societally critical domains such as recidivism prediction and loan approval.
In this work, we consider the potential bias based on protected attributes (e.g., race and gender), and
tackle this problem by learning latent representations of individuals that are statistically indistinguishable between protected groups while sufficiently preserving other information for classification.
%We formulate a ``discriminative score'' to evaluate 
% We propose that the learned latent representation of data is non-discriminative 
% if it is able to make the protected attribute statistically indistinguishable in the latent space, 
% %by invalidating classification tasks against the protected group, 
% while sufficiently preserving other information.
% We then try to find a non-discriminative representation by optimizing our proposed metric in a minimax adversarial gaming framework,
To do that, we develop a minimax adversarial framework
% which consists of
with a \textit{generator} to capture the data distribution and generate latent representations, 
and a \textit{critic} to ensure that the distributions across different protected groups are similar. 
%if the current representation is non-discriminative and provide a useful gradient for the generator. 
%an adversarial zemel2013learning to distinguish members of protected groups.  
Our framework provides theoretical guarantee with respect to statistical parity and individual fairness.
Empirical results on four real-world datasets also show that the learned representation can effectively be used for classification tasks such as credit risk prediction while obstructing information related to protected groups, especially when removing protected attributes is not sufficient for fair classification.
\end{abstract}
	
	\begin{CCSXML}
		<ccs2012>
		<concept>
		<concept_id>10002951.10003260.10003282.10003292</concept_id>
		<concept_desc>Information systems~Social networks</concept_desc>
		<concept_significance>500</concept_significance>
		</concept>
		<concept>
		<concept_id>10002951.10003227.10003351</concept_id>
		<concept_desc>Information systems~Data mining</concept_desc>
		<concept_significance>300</concept_significance>
		</concept>
		<concept>
		<concept_id>10010147.10010178</concept_id>
		<concept_desc>Computing methodologies~Artificial intelligence</concept_desc>
		<concept_significance>500</concept_significance>
		</concept>
		</ccs2012>
	\end{CCSXML}
	
	\ccsdesc[500]{Information systems~Social networks}
	\ccsdesc[300]{Information systems~Data mining}
	\ccsdesc[500]{Computing methodologies~Artificial intelligence}
	
	% We no longer use \terms command
	%\terms{Theory}
	
	\keywords{representation learning, fair learning, adversarial learning}
	
	\maketitle

\section{Introduction}

%% A quicker introduction to the background
Consequential decisions in societally critical domains, ranging from criminal justice, to banking, to medicine, are increasingly informed by predictions from machine learning models.
These machine learning models heavily rely on historical data and can inherit existing biases, leading to discriminative outcomes.
% what is the discrimination issue; some examples
% Most of the above studies have heavily relied on statistical inference and learning from historical data. However, such an approach naturally inherits the past biases and discrimination. 
For instance, in the criminal justice system, \citet{angwin2016machine} find that African-American defendants tend to be assessed with a higher risk than they actually are compared with white defendants;\footnote{This article has sparked tremendous interest, including some criticism on their methodology \citep{flores2016false,chouldechova2017fair}.} 
in online advertisting, a female user may be shown lower-priced products than a male user, even though they have similar preferences~\citep{datta2015automated,sweeney2013discrimination}. 
% Discrimination also lurks in many other domains relevant to credit lending, recruitment, college admissions, and medical diagnosis. 
As a result, fair machine learning is emerging as a central issue for deploying machine learning models in human society~\citep{holstein2018improving}.
%  \citep{corbett2018measure}.

We focus on the issue of fairness with respect to protected attributes such as gender and race in a classification setting to ensure ``fair'' decisions across protected groups.
% The definition of fairness can be extremely tricky.
% literature study: 1) the definition of non-discriminative; 2) non-discriminative approaches
However, the formulation of ``fairness'' is non-trivial and 
a growing body of research has examined a variety of definitions and developed computational approaches for achieving desired characteristics \citep{pleiss2017fairness,hardt2016equality,agarwal2018reductions,hajian2013methodology,luong2011k,calders2010classification,kamishima2012regularizer,kamiran2010discrimination,kamiran2012decision,hajian2013methodology, kamiran2012data, vzliobaite2011handling, feldman2015certifying}.
% conducted to address the discrimination issue from different perspectives. 
%considered this issue in recent years from different perspectives to yield non-discriminative results.
For example, {\em statistical parity} entails that 
% the demographics of the set of individuals receive
the proportion of the individuals in a protected group classified as positive instances are identical to the proportion of the whole population, and 
% \citet{luong2011k} and \citet{kamishima2011fairness}
\citet{kamishima2012regularizer} use regularization 
% or re-label the training data 
techniques to achieve statistical parity.
% ensure that the proportion of the individuals in the protected group classified as positive instances are identical to the proportion of the whole population. 
% Standard learning approaches either use regularizer or re-label the training data to achieve statistical parity.
Other popular metrics include false negative rates, false positive rates, and calibration within groups (see \citet{corbett2018measure} for a recent survey).
Here we highlight two important theoretical results:
1) simultaneously satisfying multiple fairness metrics can be challenging and even provably impossible for three intuitive metrics \citep{kleinberg2016inherent};
2) \citet{dwork2012fairness} show that if the distributional distance between features of different groups is small, Lipschitz conditions imply both 
\emph{statistical parity} and \emph{individual fairness}\footnote{Individuals that are similar to each other should receive similar predictions.}
% , which are two principles of evaluating fairness of classification, .

% \citep{dwork2012fairness} and \cite{zemel2013learning} further consider that any two individuals who are similar (except for the protected attribute) should be classified similarly in a particular task. 
% They both try to solve the above-mentioned problem by finding good representations for individuals. 
% \citep{pleiss2017fairness} consider a single error constraint, meaning that individuals in the protected group should have equal false-negatives rates as other groups. 
% It is important to note that .
% Consequently, our work 
% \citet{kleinberg2016inherent} formalize three different fairness constraintss and further prove that it is impossible that any single method can satisfy all these conditions simultaneously.
% Their results suggest that instead of incorporating different fairness constraints, it is better to optimize one or trade-off between them.  
We build on the theoretical results in \citet{dwork2012fairness} and propose an adversarial representation learning framework that achieves statistical parity and individual fairness.
Specifically, we formulate fairness as an optimization problem of learning representations for individuals, such that given an individual's latent representation, a task-specific classifier can obtain good performance, while one can hardly distinguish individuals in any protected groups.
% with different protected attributes. 
%another classifier can not distinguish individuals with protected attributes from others. 
% We call such representations ``fair representations'', which %aims to 
% prevent
% With the help of fair representations, it is expected that the risks of 
% both direct and indirect discrimination. % are prevented. 
% Specifically, 
Our approach reduces the Wasserstein distance between the feature distributions of people in different protected groups through an adversarial framework.
Namely, a generator learns the data distribution and transform the original (biased) features into latent representations, while a ``critic'' ensures the distributions of latent representations are indistinguishable across projected groups.
%  and provide a theoretical link between this approach and two important fairness constraints, individual fairness and statistical parity~\citep{dwork2012fairness}. 
% The model is optimized in a minimax gaming framework.
%For illustration, we visualized the original attributes and the fair representation learned by our approach in Figure~\ref{fig:pca_illustrate}. One may see that not only the distributional difference between males and females is reduced; the classification also seems to be fairer, for with original features, the classifiers are prone to classify males as having more than \$50,000 annual income than females. 
% See details in Sections~\ref{sec:setup} and Section~\ref{sec:model}. 
%\augment{Todo: highlight the difference between our work and others. Hilighted below.}
%Our work generates continuous representations of data in latent feature space, in which information of the protected attribute is effectively deprived. The representation is effective for transfer learning, and can be applied to multiple tasks. 

The advantage of our framework is twofold.
First, our framework provides theoretical guarantees on two important fairness metrics, statistical parity and individual fairness.
Such guarantees are particularly strong given the impossibility results in \citet{kleinberg2016inherent}. 
%Second, different from using an adversarial framework for end-to-end learning in \citet{zhang2018mitigating}, our learned ``fair'' representations can be used by downstream machine learning models and these fairness properties can  still be maintained with reasonable assumptions because the information related to protected attributes have been effectively obstructed.
Second, compared to prior work on adversarial learning for fair classification \citep{edwards2015censoring,zhang2018mitigating,madras2018learning}, our framework directly minimizes the distributional distance of latent representations between protected groups rather than minimizing the ability to predict protected attributes. 
% which is different from existing works, such as \citet{edwards2015censoring,zhang2018mitigating,madras2018learning}. 
Our method effectively blocks the information related to protected attributes and hence ensures fairness properties with any downstream models that satisfy Lipschitz conditions.
% \reminder{I rewrote the second according to my understanding.}

% Notice that we cannot simply remove the protected attribute from the data to avoid discrimination, due to the existence of \emph{indirect discrimination}. 
% %One important aspect of non-discriminatory method is pre-processing the data, i.e. modifying the data such that no discriminatory result may be obtained by downstream data mining tasks. 
% %However, 
% %One should note that 
% In particular, the protected attribute may contribute to discrimination either directly or indirectly~\cite{zliobaite2015survey,hajian2013methodology}. When the protected attribute is presence, it may influence the classification results 
% %data mining process 
% directly; but when it is absence, it may still exert influence through other variables related to it. 
%The existence of \emph{indirect discrimination} is why we cannot simply remove the protected attribute from the data. 

%Please notice that, in this paper, we focus on studying the cases involved with only one protected attribute. We leave the extension of our approach to more protected attributes as the future work.   

% our work

\hide{
to learn representations of data that make it easier to extract useful information when building classifiers or other predictors
}

 \begin{figure}[t]
	\centering
	\subfigure[Original features.]{							   \includegraphics[width=0.28\linewidth]{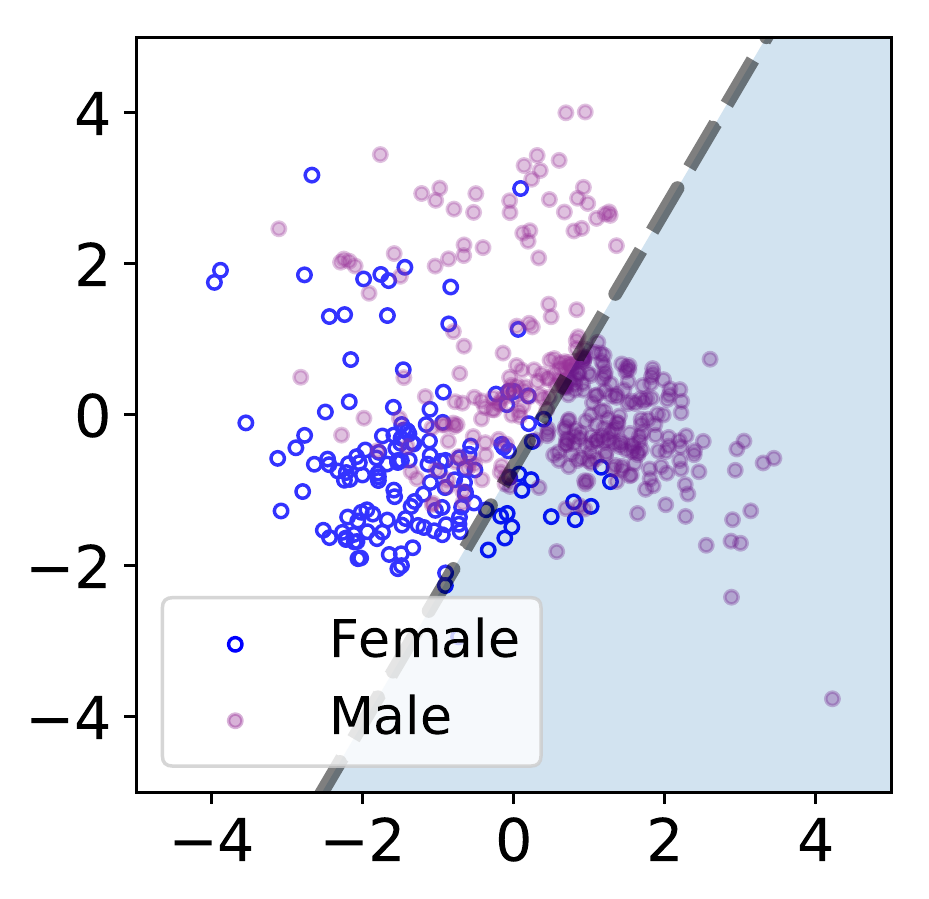}
		\label{fig:pca_illustrate:a}
	}
	\subfigure[Original features except gender.]{							   \includegraphics[width=0.28\linewidth]{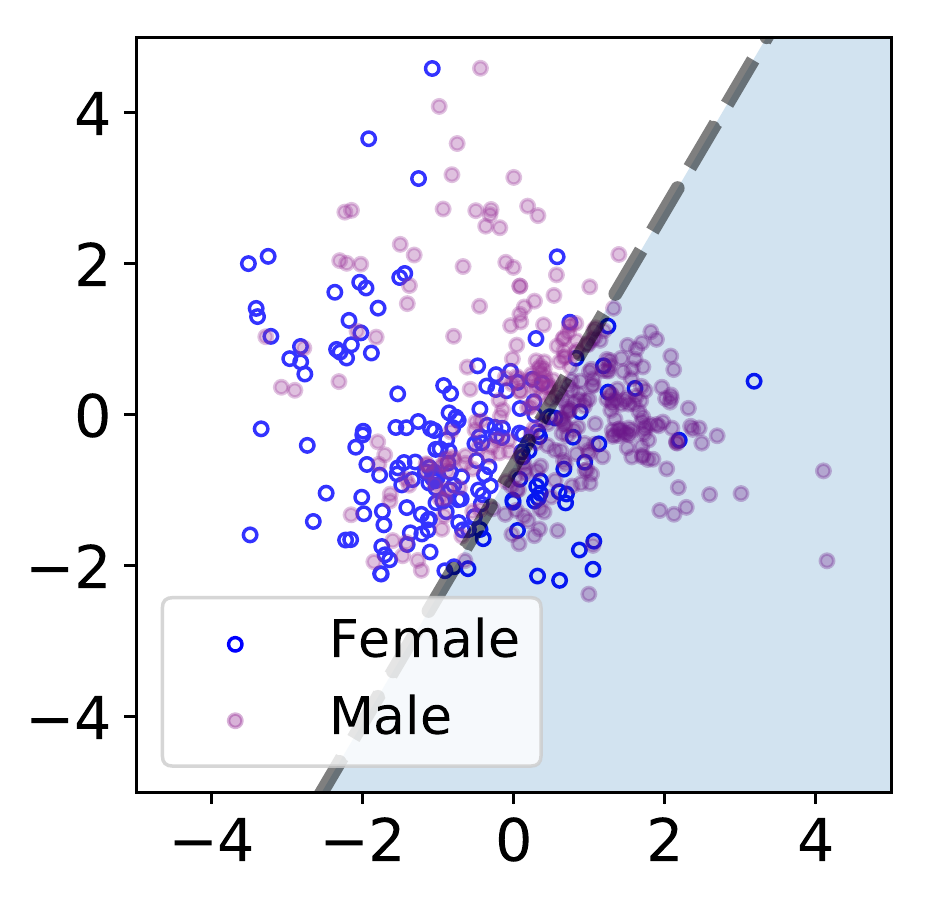}
		\label{fig:pca_illustrate:b}
	}
	\subfigure[Fair Representation.] {
		\includegraphics[width=0.32\linewidth]{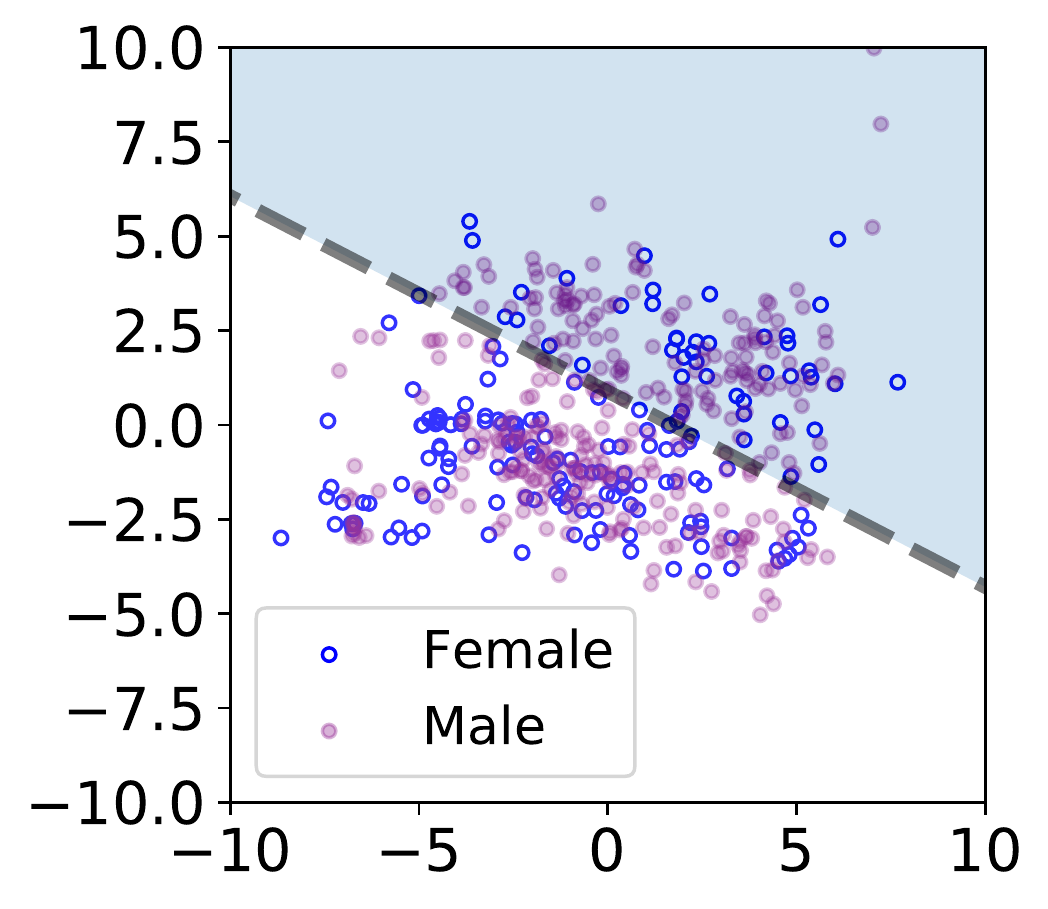}
		\label{fig:pca_illustrate:c}
	}
	\caption{Illustration of fair representations.  
	% \small 
	Figure~\ref{fig:pca_illustrate:a} shows the original features, Figure~\ref{fig:pca_illustrate:b} shows the the original features excluding the gender attribute, and Figure~\ref{fig:pca_illustrate:c} shows the fair representations learned by our model. 
	%Our model attempts to learn a representation in which males and females has no statistical difference. 
	The features are scaled and projected into \(\RR^2\) using PCA. The solid orange points represent male individuals, and hallow blue points represent females. The solid lines are decision boundaries by linear classifiers, where the light blue areas are classified as having more than \$50,000 annual income, and the white areas are the opposite. 2,000 random samples from the Adult dataset are plotted.
		%We managed to reduce the distributional difference between the male's group and the female's group, and thus produced fair classification results satisfying both statistical parity and individual fairness. 
	}
	\label{fig:pca_illustrate}
	%\vspace{-0.1in}	
\end{figure}

To illustrate the intuition of our fair representations, we present an example from the Adult dataset~\cite{Adult:2013}, where individuals are labeled by whether their income exceeds \$50,000 per year. 
We visualize the dataset by projecting the attributes to \(\RR^2\) by PCA in Figure~\ref{fig:pca_illustrate}. The black dashed line is the decision boundary of logistic regression w.r.t. the label, and different colors indicate different gender. 
In this dataset, females are less likely to have a high income compared to their male counterparts, and such bias carries over into the predictions from the original features. 
Based on the original features, the distributional difference between males and females is clear (Figure~\ref{fig:pca_illustrate:a}) and 
% our experiments show that 
the logistic regression classifier gives males about 2.39 (Figure~\ref{fig:exp:fair:cls}) more chances of having a high income. 
% What immediately comes to mind
A straightforward strategy to mitigate such bias is to remove the attribute ``gender'' from the data. However, even after removing ``gender'', the distributional difference still exists in Figure~\ref{fig:pca_illustrate:b}, and males are now 1.95 (Figure~\ref{fig:exp:fair:cls}) more likely to have a high income. This is because the data suffers from ``indirect discrimination''~\cite{zliobaite2015survey,hajian2013methodology}: other attributes relevant to ``gender'' may still inflict discrimination upon females. 
This motivates us to reduce both direct and indirect discrimination by learning fair representations of data that are independent of gender. Figure~\ref{fig:pca_illustrate:c} showcases our learned fair representations, where one can hardly distinguish males and females as a whole, and males and females are equally likely to have a high income when this representation is used by a learning algorithm. 
% \chenhao{what is the accuracy difference of the new representation?}

To further evaluate the effectiveness of our algorithm in promoting fair decision-making, we conduct experiments on four real-world datasets in Section~\ref{sec:exp}. 
First, we use {\em gender} as the protected attribute for all four datasets. 
Experimental results show that classification is fairer using our learned representations than by using original attributes, with or without the protected attribute, in terms of multiple fairness metrics.
% both statistical parity and consistency. 
% \hide{
% \reminder{Added description based on our new experiments.}
% Furthermore, we investigated how the model actually works by looking into how the model uses the original features. 
% We also checked whether the model may help actually discover and mitigate the effect of discrimination manually installed on the data by flipping some female's labels. This suggests that our model may also be a tool for discrimination discovery.
% Finally, }
Furthermore, we extended our model to the case where the protected attribute is categorical with more than two classes, e.g., {\em race} and demonstrate how our method can also be used to discover biases.
% ``race'' as an example.
% On PPDai's Fraud dataset, simply taking out the protected attribute yields fairer classification than our method and LFR~\citep{zemel2013learning}. 
% The experimental results are discussed in further details in Section~\ref{sec:exp}. 
% \augment{I modified here.}

% We then consider past relevant work in Section~\ref{sec:related} and conclude our work in Section~\ref{sec:conclusion}. 

Finally, we highlight our contributions as follows:

\begin{itemize}[topsep=2pt, itemsep=0in,leftmargin=*]
% \item We propose a new theoretical metrics to measure the fairness of a prediction model, by both ensuring the model performance on the prediction task and hiding the protected attributes. \augment{imprecise}
% \item We propse to utilize propensity score matching as a theoretical metric to measure the fairness of a prediction model.
%\item We propose a new theoretical metric to measure the fairness of a learned representation, that all classifiers against the protected attribute are invalidated. We establish its relation between $t$-closeness principle, which calls that the distribution of the protected group and the overall population be similar.
%\augment{You may want to examine this, I'm not sure about it.}
\item We use the Wasserstein Distance to measure if a learned representation is fair and as a result, provide theoretical guarantee
% provide a theoretical framework 
for two common formulations of fairness constraints, statistical parity and individual fairness. 
% by comparing the distributions over individuals with different protected attributes in the latent space. 
%The Wasserstein distance is used in our metric,  
\item We develop an adversarial framework to incorporate our new metric and learn fair representations for given individuals. % and formulate the fairness optimization problem into a representation learning task. 
\item We validate our proposed framework by conducting extensive experiments on real-world datasets. Using the learned latent representations, we can obtain a competitive prediction performance while ensuring that the protected attributes are unidentifiable. 
We will release our code upon publication.
%letting the population with the protected attribute indistinguishable from others. 
\end{itemize}

	%\input{sketch}
	%!TEX root = main.tex

\section{Related Work}
\label{sec:related}
%\augment{somewhere, elaborate on that "removing the protected attribute is not enough". -> more systematic approach to fairness.} 
\hide{
\vpara{Summary.}
In this part we introduce previous research relevant to our study. First, we summarize works that provided theoretical ground and motivation for the study of fair learning and techniques to discover discriminatory patterns. Then we discuss machine learning methods to ensure fair results. After that, we review recent works on fair representation learning.
Finally, we list recent progress in other related fields, including privacy preservation, adversarial learning, and autoencoder. 
}

We summarize related work in the following three strands.
% fair representation learning, adversarial learning, and .

\vpara{Fair representation learning.}
Most relevant to our work are studies on fair representation learning.
% In recent years, many works have been dedicated to finding intermediate fair representations of data. Closely related to our work, \citet{dwork2012fairness} noted that Lipschitz condition implies statistical parity if and only if the Wasserstein Distance between two groups is small. 
\citet{zemel2013learning} was the first to propose learning fair intermediate representations: their method involves finding \(k\) prototypes in the same space as the input data. Their approach is similar \(k\)-means, which assigns each sample point to the closest prototype while adding the fairness constraint and classification performance to the optimization objective. However, their representation is inflexible and loses too much information.

Recent advances in generative adversarial networks have also inspired studies that learn fair representations~\citep{edwards2015censoring,beutel2017adv,madras2018learning}. \citet{edwards2015censoring} first made the connection between adversarial and fair machine learning, and \citet{beutel2017adv} analyzed \citeauthor{edwards2015censoring}'s algorithm and found that their method can be instable when the demographics of the protected attribute are imbalanced. \citet{zhang2018mitigating} used a adversarial agent which attempts to predict the protected attribute solely based on the classifier output; \citet{madras2018learning} used an adversary objective based on the learned representations in order to achieve statistical parity and equalized odds. These methods are all based on using a classifier that predicts the protected attribute as the adversarial component. 
% \chenhao{it really begs the question why these are not used as baselines ... but I guess that we cannot do anything about it.}

However, 
% using a classifier as an adversary could be troublesome, since to achieve fairness, it 
such an adversarial setup requires that 
\emph{any} adversary cannot predict the protected attribute, which can be too difficult to optimize in practice.
%  to optimize. 
% In practice, a sufficiently strong adversary is required, and it tunning of the architectures of the adversary and the autoencoder to achieve balance is tricky. Even when the balance is achieved, one cannot be sure if a different type of classifier can predict the protected attribute.
Our method directly reduces the distributional distances between different groups induced by the protected attribute in the latent space. Thus, it is automatically ensured that \emph{any} classifier cannot predict the protected attribute better than random guessing and the two fairness constraints are satisfied. Hence, with our model, a rather simple architecture is enough for both efficiently preserving information and ensuring fairness constraints, which makes the optimization of our model much easier. 

\vpara{Adversarial learning and autoencoder.}
% The framework of autoencoder networks is summarized in 
\citet{wang2014auto} summarizes the framework of autoencoders, and there exists many variations 
% are proposed, such as in
~\citep{bengio2013denoising,kingma2013variational}.
Adversarial networks were first proposed by \citet{goodfellow2014gan}. 
% \citet{yu2016seqgan} extended the original model architecture to apply to sequences. 
Most relevant to our work is WGAN proposed by \citet{arjovsky2017wgan}, % proposed WGAN
% , 
which replaces the KL-divergence with the Wasserstein Distance, to improve stability of convergence.

\vpara{Discovering biases.}
% One should first know where and how discrimination exists in data, and the field of study focused on this is called \emph{discrimination discovery}.
Another important direction we have not discussed is to discover biases in algorithmic systems. 
Many studies give concrete examples and case analysis of biases in algorithmic systems~\cite{sweeney2013discrimination,datta2015automated,roth1997effects,kamiran2012classifying,romei2013discrimination,mikians2012detecting}. 
The existence and origin of algorithmic bias and its legal background were thoroughly surveyed by~\citet{zliobaite2013algorithmicbias,romei2013discriminationsurvey,gellert2013antidiscrimination}. 
Several criteria are proposed for quantifying the extent of biases%
% , % such as in the works by
~\citep{luong2011k,dwork2012fairness,zemel2013learning,romei2013discriminationsurvey,pedreschi2012survey,pedreschi2009measure}.

\section{Problem Definition}
\label{sec:setup}
%In this section, we give formal definition of \textit{fair representation}, 
%In this section, we formulate the problem of learning fair representations. 
%Generally, we define an individual's \textit{fair representation} as the one that loses information about the protected attributes while preserving as much other information of that individual as possible. 
%Before we go through the details, we first give necessary definitions and notations. 

%\subsection{Preliminaries}
%\vpara{Data representation.} 
Given a set of individuals $\{u_1, \cdots, u_N\}$, each individual $u_i$ is denoted as a triple $(\mathbf{x}_i,  y_i, p_i)$, where \(\mathbf{x}_i \in \RR^\nattrs\) represents their attributes, $y_i \in \{0, 1\}$ is the label used for the specific relevant classification task, and $p_i \in \{0, 1\}$ is the \textit{protected attribute} that we hope to conceal in the classification process. 
%In particular, each vector $\mathbf{X}_k \in \mathbb{R}^\nattrs$ represents attributes of the individual $i$ and each vector $\mathbf{X}_i$ has $\nattrs$ attributes. We may also write the samples in the matrix form: \(\mathbf{X}=[\mathbf{X}^T_1 \ldots \mathbf{X}^T_\nattrs]^T\).
%Each individual $i$ is assigned with a discrete valued label $Y_i$, indicating some property we are concerned about.

%\vpara{Protected attribute.}
%Each individual is also assigned another discrete variable \(P_i\), assigning the individual an identity which we hope to conceal to the decision-making process. 
Indeed, there are some attributes, such as gender, race, age, etc., which should not be used in the decision-making process, because using it would violate fairness and legitimacy of the result, no matter how accurate the model is fitted on the observed data. 
%Among the $d$ attributes of data, there are some which, when being used in the decision-making process, would violate fairness and legitimacy of the result, no matter how accurate it is on the training and testing data. 
%Such attributes are refereed to as the \emph{protected attributes}, such as gender, race, age, etc.. 
%However, protected attributes are not supposed to have no impact in the decision-making process, which should be made out of considerations of an individual's traits and talents, instead of a irrelevant group to which she belongs.
It is our intention that the protected attributes have no impact on the decision-making process. This process should give due consideration to an individual's traits and talents instead of a relevant group to which they belong. To this end, simply ignoring the protected attribute alone is not enough, because other attributes are often correlated with the protected attribute, and it is often useful when removing possible bias.   

In this section, we consider only binary protected attributes. In Section~\ref{sec:multiclass}, we extended the model to the multi-class scenario and conducted experiments.  
As such, samples can be categorized into different \textit{groups} according to their protected attributes. 
Groups that suffer from discrimination are called \emph{protected groups}.
% ~\footnote{For notational convenience and to avoid ambiguity, we sometimes use \(\{X^{(i)}_k\}_{1\leq k \leq \nsamps_i}\) to represent samples for which \(P_k=i\).}. 
Furthermore, we define \(\nsamps_c\) as the number of samples that belongs to a particular protected category (i.e., \(p_i=c\)) and denote 
% \(X^{(c)}_j, j=1,\ldots,\nsamps_c\)
\(U^{(c)}\)
as the subset with \(p_i=c\).

It is natural to assume that some different distributional rules govern the generation of data of the different groups. 
We define two probability measures over the Borel sets of \(\RR^d\), say \(\mu_0\) and \(\mu_1\), as the generator of data with \(p_i=0\) and \(p_i=1\), respectively. 
%Their distributions are written as \(F_0\) and \(F_1\). 

%\vpara{Classification tasks.}
%In machine decision-making settings, classification tasks are often involved. Here, we define a classifier as a mapping \(\Psi:\RR^\nattrs\rightarrow[0,1]\), which assigns risk scores for samples in \(\RR^\nattrs\). Fairness here is therefore concerned with the classification results; it should be ``fair'' for different groups of the protected attribute. We shall introduce the definition of fairness here immediately. 

\vpara{Learning fair representations.} 
The goal is to learn a representation vector for each individual in some latent feature space $\RR^\ndim (\ndim\leq \nattrs)$. 
We define a mapping $f:\RR^\nattrs\rightarrow \RR^\ndim, \mathbf{x}_i \mapsto \mathbf{z}_i$, where $\mathbf{z}_i$ is the desired representation of $\mathbf{x}_i$. 
Each component of $\mathbf{z}_i$ is real-valued and continuous.
Let $\mathbf{Z} = [\mathbf{z}_1, \ldots, \mathbf{z}_\nsamps]$, for a classification task, we first map $\mathbf{X} = [\mathbf{x}_1, \ldots, \mathbf{x}_\nsamps]$ to $\mathbf{Z}$, and then map $\mathbf{Z}$ to $\mathbf{Y} = [y_1, \ldots, y_\nsamps]$. 

Usually, a good representation vector is expected to preserve most of the information from the original vector. 
However, when fairness is at risk, we wish to further impose some constraints on the distribution of instances in the feature space, such that statistical parity and individual fairness are met.

\vpara{Fairness constraints.} 
We define \(\Psi:\RR^\ndim\rightarrow[0,1]\) as the classifier that assigns each individual a classification score. 
In this paper, we discuss two popular realizations of fairness constraints, as listed below. Similar ideas are also discussed in~\citet{dwork2012fairness,corbett2017fairness,zemel2013learning}.

\begin{itemize}
	\item \emph{Statistical parity}~(\citet{corbett2017fairness}): all groups with the protected attribute receive similar expected classification scores, i.e.,  $\expt[\Psi(\mathbf{Z})|P] = \expt[\Psi(\mathbf{Z})]$;
	\item \emph{Individual fairness}~(\citet{dwork2012fairness}): similar individuals receive similar scores, i.e., $\vert\Psi(\mathbf{z}_1)-\Psi(\mathbf{z}_2)\vert \leq K \Vert \mathbf{z}_1-\mathbf{z}_2\Vert_2$.  In other words, the difference between the classification scores is bounded by the difference between individual features. 
\end{itemize}

These are two representative definitions of algorithmic fairness from relevant literature. Throughout the paper, we use these definitions to refer to their respective notions of fairness.

%Suppose that \(\Psi:\RR^\nattrs\rightarrow[0,1]\) is a classifier that assigns each individual a classification scores. Then statistical parity dictates that \[\expt[\Psi(\mathbf{X})|P] = \expt[\Psi(\mathbf{X})]\]
%And individual fairness can be formulated as
% \[\vert\Psi(\mathbf{x})-\Psi(\mathbf{y})\vert \leq K \Vert \mathbf{x}-\mathbf{y}\Vert_2\]
% That is, the risk scores difference is bounded by the difference of individual features. 

%Inspired by differential privacy and t-closeness, a fair representation should be able to encode as many information about the original data, while obstructing the identification of protected attributes. 

\hide{
Generally, we define an individual's \textit{fair representations} as ones that lose information about the protected attributes while preserve as much other information of that individual as possible. 
In other words, 
%In other words, it should has the following two characters: 

\begin{itemize}
\item It contains no or little information about protected attributes. So one can hardly distinguish the protected groups from others according to fair representations.
\item Still, it can preserve as much information, except the protected attribute, as possible.   
\end{itemize}

 }

\section{Our Approach}
\label{sec:model}
In this section, we propose \textit{a minimax adversarial gaming framework} to learn the fair representations of the relevant individuals. 
%Generally, we define an individual's 
Intuitively, to  meet fairness constants, the \textit{fair representations} are those that lose the information about the protected attributes while preserving as much of the other information of that individual as possible. 
In other words, 
%In other words, it should has the following two characters: 
\begin{itemize}
\item It contains little or no information about the protected attributes. Therefore, it is difficult to distinguish the protected groups from the others according to fair representations.
\item Still, it can preserve as much information as possible except for the protected attribute. \textsf{, except for the protected attribute, as possible. }  
\end{itemize}
\subsection{Framework}
%We will introduce how to quantify the above idea in the next section.  
To hide the protected attributes, we impose the constraint that the conditional distributions given the protected attributes are identical across the feature space. 
Therefore, we define a \textit{fairness metric} to evaluate the quality of a learned representation: 
\begin{equation}
    \label{eqn:fairness}
    L = L_A + \alpha L_D 
\end{equation}
The proposed metric $L$ comprises of two parts:
%Here, $f$ is the embedding mapping, and 
$L_A$ is the information loss term, evaluating how much information the embedding mapping $f$ has not preserved compared with the original feature matrix; 
%$d(\cdot, \cdot)$ is some functional which measures the distances between two probability measures; 
%and \(L_D =d(\Phi(\mu_0), \Phi(\mu_1)) \). 
and $L_D$ measures how close the distributions over different groups of the protected attribute (i.e., $\mu_0$ and $\mu_1$) are in the latent space. 
We expect that a fair representation has small-valued $L_A$ and \(L_D\).
The hyperparameter \(\alpha\) controls the trade-off of the system.

Later in this section, we introduce the relationship between the proposed fair metric $L$ and the fairness constraints theoretically.
Prior to that, we present the overall structure of our approach.  
 \begin{figure}[t]
 \centering
   \includegraphics[width=0.25\textwidth]{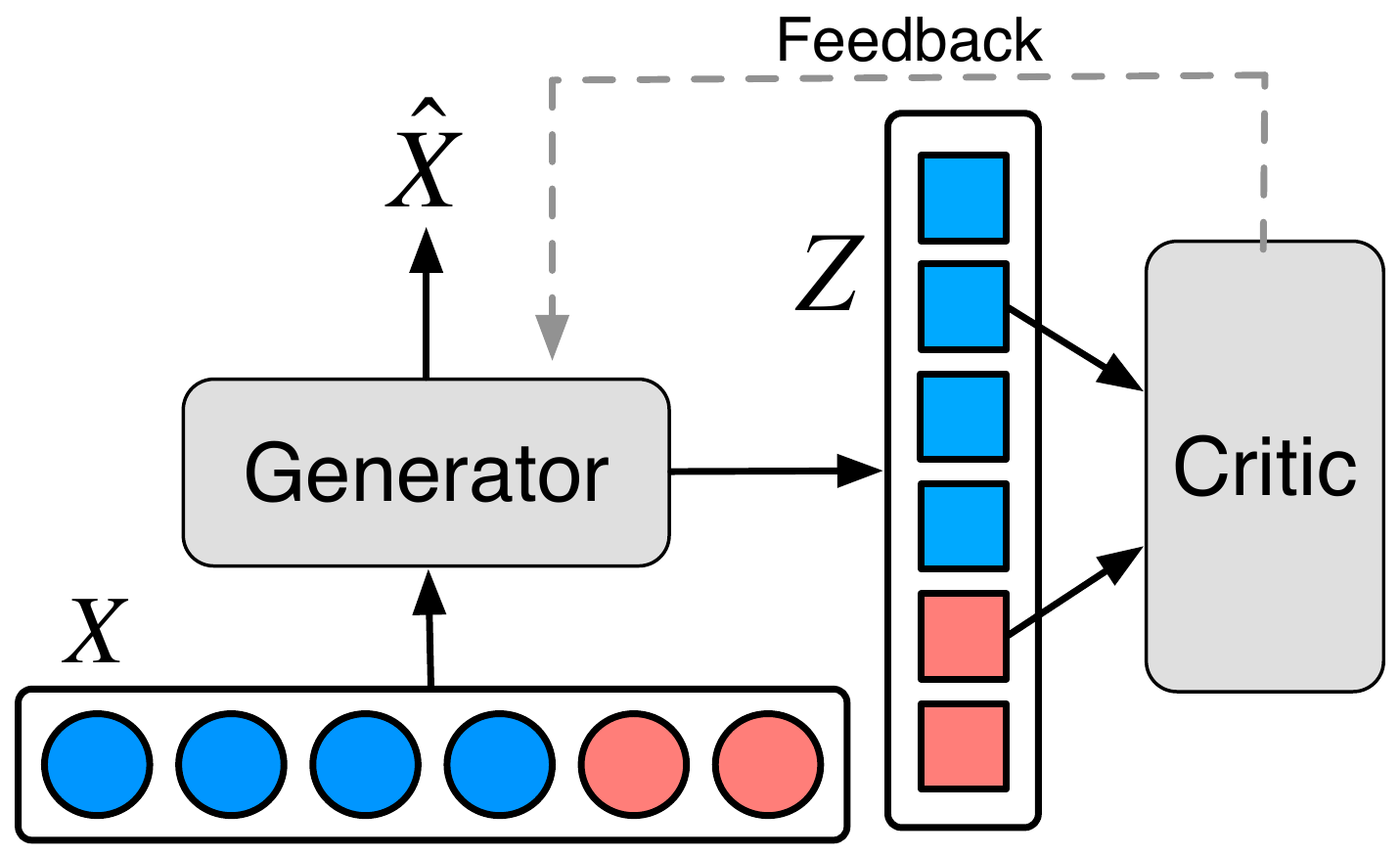}
   %\vspace{-0.1in}
   \caption{Structure of our proposed model. 
   % \small 
    Red circles (square) represent $\mathbf{X}$ ($\mathbf{U}$) of individuals in protected group, while blue ones indicate that of other individuals. 
    \normalsize 
%    \chenhao{This figure needs to be updated with the new notation}
   %, which consists of an autoencoder to learn latent representation for an individual, and a zemel2013learning to determine the value of the protected attribute according to the individual's representation vector. 
   }
   \label{fig:framework}
 %\vspace{-0.1in}	
 \end{figure}
Overall, as Figure~\ref{fig:framework} shows, the proposed framework consists of two components: 1) the generator, which takes individual features $\mathbf{X}$ as the input, learns representations $\mathbf{Z}$ of individuals, and recovers individual features according to the learned representations and 2) the critic, which measures the distance between $\mu_0$ and $\mu_1$. 
The critic guides the learning process of the generator, ensuring that the different values of the protected attributes can hardly be distinguished in the representation space while the generator provides the representations of individuals. 
Note that the goal of the critic and the generator correspond to $L_D$ and $L_A$ respectively. 
%\yang{add more descriptions about the whole framework, describe generator and critic, and maybe update fig 1}
We next introduce how to implement the critic and the generator. 

\vpara{Generator.} 
The generator aims to generate data representations. 
%\vpara{Generator: generating data representations.} 
In this paper, we use an autoencoder as an example to implement the \textit{generator} and learn the representations of the individuals.  
In particular, the autoencoder comprises of two multilayer perceptrons: the encoder $f$, which takes data as input and produces latent representation; and the decoder $g$, which operates inversely. 
We define $L_A$ in \eqref{eqn:fairness} as the mean square error between $X$ and $g(f(X))$, since $g\circ f$ tries to reproduce $X$ when $X$ is given. 

For simplicity and interpretability, we use a single linear layer for both the encoder and the decoder. 
Thus, the latent features are simply linear transformations of the original attributes. 
We denote, therefore, \(\mathbf{Z} = \mathbf{X}\mathbf{A}\), where \(\mathbf{A}\in\RR^{\nattrs\times\ndim}\) defines a linear mapping to the feature space. 
%The critic is also a linear mapping, and is therefore a vector in \(\RR^{\ndim}\).

\(\mu_0, \mu_1\) and the encoder \(f\) induces probability measures on the latent space \(\RR^\ndim\), \(f(\mu_0)\) and \(f(\mu_1)\), representing the distributions of data with different \(p_i\)'s in the latent space.
Since we are always concerned with the distributions on the latent space, for notational convenience, in the following paper, we use \(\mu_0, \mu_1\) themselves to refer to the probability measures on the latent space without specifying \(f\). 

\vpara{Critic.}
The goal of the critic is to keep the protected attributes similar for different groups. 
%\subsection{Critic: Keeping Protected Attributes Similar for Different Groups}
To calculate $L_D$, we measure the distance between $\mu_0$ and $\mu_1$ by the Wasserstein Distance, or the Earth Mover's Distance (EMD), as follows: 
\begin{equation}
L_D=D_W(\mu_0, \mu_1) = \inf_{\gamma\in\Gamma(\mu_0, \mu_1)}\left(\int_{\RR^\ndim\times\RR^\ndim} \Vert \mathbf{z}_0-\mathbf{z}_1 \Vert_2 d\gamma(\mathbf{z}_0,\mathbf{z}_1) \right)
\label{eqn:emd}
\end{equation}
The sole context is Euclidean metric space. 
\(\mu_0, \mu_1\) are two probability measures defined on \(\RR^\ndim\), which represent the distribution of data with different \(p_i\) in the latent space. \(\Gamma(\mu_0, \mu_1)\) is the collection of all probability measures on \(\RR^\ndim\times\RR^\ndim \), with two marginal distributions of \(\mu_0\) and \(\mu_1\) respectively. 

This definition is based on the intuition of optimal transport, i.e., the minimum ``weight'' that needs to be transferred to transform the density of \(\mu_0\) into \(\mu_1\). 

Unfortunately, the computation of \eqref{eqn:emd} is intractable. 
Instead, we use the Kantorovich-Rubinstein Duality~\citep{villani2008optimal}:
\begin{equation}
D_W(\mu_0, \mu_1) = \frac{1}{K}\sup_{\Vert D \Vert_L \leq K} \expt_{\mathbf{z}\sim\mu_0}D(\mathbf{z}) - \expt_{\mathbf{z}\sim\mu_1} D(\mathbf{z})
\label{eqn:emd_kr}
\end{equation}
Here, \(D\) is our \textit{critic}, which is taken across all functions \(\RR^\ndim\rightarrow\RR\) satisfying the Lipschitz condition. The Lipschitz condition provides that \(\vert D(\mathbf{z}_0)-D(\mathbf{z}_1)\vert \leq K\Vert \mathbf{z}_0-\mathbf{z}_1\Vert_2 \). $K$ here is called the Lipschitz constant, denoted by \(\Vert D\Vert_L\), as in \eqref{eqn:emd_kr}.

In this paper, we use a linear mapping to implement \textit{the critic} \(D\), which is used to approximate the EMD. This approach, in turn, provides a useful gradient for the generator to ensure that the EMD is close to zero. 
The efficiency of this approach is tested in~\citep{arjovsky2017wgan}. 
%\yang{I think we need to cite WGAN somehere, and I did it here.}
Therefore, our objective can be written in the following minimax form: 
\small
\begin{eqnarray}
\label{eqn:model:loss_func}
\min_A{}\max_D{} L(A,D) & = & L_A + \alpha L_D  
\end{eqnarray}
\normalsize
Here, \(L_A\) is the information loss term, and \(L_D\) is the approximation of EMD provided by the critic \(D\). \(L_D\) is first maximized to ensure good approximation and to yield a usable gradient for the generator. Then the generator \(A\) minimizes both the information loss and the EMD distance. 
Note that the critic is a real-valued linear mapping, and is therefore a vector in \(\RR^{\ndim}\).

%\vpara{Imposing Lipschitz Constraint}
To ensure Lipschitz condition in general multilayer neural networks, we may clamp the model parameters to the range of \([-c,c]\), where \(c\) is a positive real number. This is the approach practiced by~\cite{arjovsky2017wgan}. 
Since neural networks are differentiable, this ensures bounded first-order derivative and therefore the Lipschitz condition. 

\subsection{Relation to Fairness Constraints}
\citet{dwork2012fairness} explained that statistical parity and individual fairness can be jointly achieved if the Wasserstein Distance between two groups is small. 
%\yang{They mentioned this, I think we need to cite this too. }
Here, we give concrete reasons for why we have selected the Wasserstein Distance as an optimization objective by showing that it may ensure the fairness of the classification results on the learned features for a wide range of well-conditioned classifiers. 

\begin{definition}
	Suppose \(\Psi\) is a mapping \(\RR^\ndim\rightarrow\RR\) satisfying 
	\[
	|\Psi(\mathbf{z}_0)-\Psi(\mathbf{z}_1)|\leq K \Vert \mathbf{z}_0 - \mathbf{z}_1\Vert_2
	\]
	where \(K\) is a positive constant, then we say \(\Psi\) satisfies the Lipschitz condition, and denote it by \(\Vert \Psi\Vert_L \leq K\), where \(K\) is called the Lipschitz constant. 
	\label{def:lipschitz}
\end{definition}

\begin{theorem}
	If \(\Psi\) is a mapping \(\RR^\ndim\rightarrow\RR\) representing a classifier, 
	and \(\Vert \Psi\Vert_L\leq K\), \(\mu_0, \mu_1\) are two probability measures defined over \(\RR^\ndim\) that generates the two categories of the protected attribute, then the individual fairness is bounded by the Lipschitz constant as is the Definition~\ref{def:lipschitz}. Furthermore, the statistical parity is bounded by the Wasserstein disance and the Lipschitz constant:
	\begin{equation}
		\vert\expt_{\mathbf{z}\sim\mu_0}\Psi(\mathbf{z})-\expt_{\mathbf{z}\sim\mu_1}\Psi(\mathbf{z})\vert\leq KD_W(\mu_0,\mu_1) 
		\label{eqn:fair_bound}
	\end{equation}
	\label{lem:emd_fairness}
\end{theorem}
Theorem~\ref{lem:emd_fairness} shows how to principally ensure both statistical parity and individual fairness. Clamping parameters or \(\ell_2\) weight decay both reduce the Lipschitz constant of the classifier. (\ref{eqn:fair_bound}) shows that by reducing the EMD distance, enforcing individual fairness achieves statistical parity as well.

\begin{algorithm}[t]
	\caption{The learning algorithm.}
	\label{alg:train}
    \small
	\begin{algorithmic}[1]
	\STATE Normalize the input data
	\FOR{\(n_{iter}\) training epochs}
		\STATE Sample a batch of data \(x_0^{(i)},\ldots,x_L^{(i)}\) from \(U^{(i)}\) for \(i=0,1\)
		\WHILE{not converged}
		\STATE Update the parameters of the critic:
			\[L_D = \frac{1}{L}\sum_{i=1}^L{D(f(x_i^{(0)}))-D(f(x_i^{(1)}))}\]
			\[\theta_D = \theta_D + \mu \frac{\partial}{\partial \theta_D} L_D \]
			\STATE Clipping the values of \(\theta_D\) to the range of \([-c,c]\)
		\ENDWHILE
		\STATE Update the parameters of the autoencoder:
			\[L_A = \frac{1}{L}\sum_{i=1}^2\sum_{j=1}^L (g(f(x_j^{(i)})) - x_j^{(i)}) \]
			\[L_D = \frac{1}{L}\sum_{i=1}^L{D(f(x_i^{(0)}))-D(f(x_i^{(1)}))} \] 
			\[ \theta_A = \theta_A - \mu \frac{\partial}{\partial \theta_A} (L_A + \alpha L_D) \]
	\ENDFOR
	\end{algorithmic}
\end{algorithm}

\subsection{Model Learning} 
By putting everything together, we obtain our model with the dynamics as a minimax adversarial game:  

\small
\begin{eqnarray}
    \label{eqn:model:loss_func}
    \min_A{}\max_D{} L(A,D) & = & L_A + \alpha L_D \nonumber \\
    & = & \frac{1}{\nsamps}\sum_{i=1}^\nsamps (g(f(x_{i})) - x_{i})^2 \nonumber \\
    & + & \frac{1}{\nsamps_0\nsamps_1}\sum_{i}\sum_{j} D(f(x_{i}^{(0)}))-D(f(x^{(1)}_{j})) \nonumber
\end{eqnarray}
\normalsize

Let $\theta_A$ and $\theta_D$ indicate the parameters of the autoencoder and the critic respectively, we introduce how to learn $\theta_A$, $\theta_D$, and the latent representation $\mathbf{U}$ simultaneously. Algorithm \ref{alg:train} is the training process. First, the critic is trained sufficiently until the terminal condition is met. In our experiments, we keep training the critic until the change in value is less than \(1e-3\).
Then, the autoencoder is trained to minimize the MSE loss and the EMD by using the gradients provided by the critic. 

\hide{
\subsection{Regularization}
Enforcing small weights and sparisty would increase performance of EMD reduction for both plain autoencoder and our NRL. 
In additition to \(L_A\) and \(L_D\), we also punish the encoder by the \(\ell_{2,1}\)-norm of of the weight matrix, defined as follows:
\[
\Vert \mathbf{A} \Vert_{2,1} = \sum_i \left(\sum_j \left\vert A_{ij} \right\vert^2\right) ^ {1/2}
\]
That is, the sum of the \(\ell_2\)-norm of the row vectors of \(\mathbf{A}\). The idea behind this is that some original attributes are punished such that they do not contribute to any of the components in \(\mathbf{U}\). 
}

% \subsection{Extending to the Multivariate Scenario}
\subsection{Beyond Binary Protected Attributes}
\label{sec:multiclass}
Previously, we assume 
%In previous sections, we theoretically analyzed the use of Wasserstein distance and an adversarial generative framework in learning fair representation, when 
the protected attribute is binary, i.e. gender. 
In practice, protected attributes are often multi-class categorical variables. Suppose the protected attribute is converted to numeric representation \(P=0,1,\ldots,n_p-1\), where \(n_p\) is the number of categories that the protected attribute may assume. 

%\reminder{Is the presentation of the ``constrained objectives'' here necessary? Maybe we delete it. }
Our original formulation of the model in the binary case may be viewed as an approximation of the following objective:
\begin{eqnarray}
	\min & \frac{1}{n}\sum_i (g(f(x_i)) - x_i)^2  \\ 
	\operatorname{s.t.} &   D_W(f(U^{(0)}), f(U^{(1)}))  = 0 \nonumber
\end{eqnarray}
\noindent where \(U^{(j)}\) represents the collection of samples which satisfies \(p_i=j\), \(j=0,1\).

The direct extension to the multi-class case would be adding as many constraints as to ensure that the distributional difference between any two classes is zero. However, this would add \(\binom{p}{2}\) constraints.
Therefore, it is more desirable to define the multi-class objective in a ``one-vs-rest'' manner. We denote \(U^{(-j)}\) as all the samples that \(p_i\neq j\). 
The multi-class objective is then 
\begin{eqnarray}
\min & \frac{1}{n}\sum_i (g(f(x_i)) - x_i)^2 &  \\ 
\operatorname{s.t.} &   D_W(f(U^{(j)}), f(U^{(-j)}))  = 0, & j=0,\ldots,n_p-1 \nonumber 
\end{eqnarray}
Intuitively, this adds \(p\) constraints and \(p\) critics that are needed to approximate the Wasserstein distances. However, we discovered in our experiments that using only one linear critic for all the distributions has similar performance comparing with using one linear critic for each of the distributions. Therefore, we used only one linear critic to approximate the distributions. 

\vpara{Implementation note.} 
For a given \(i\), a training iteration is exactly like Algorithm~\ref{alg:train}, replacing \(U^{(0)}, U^{(1)}\) with \(U^{(j)}, U^{(-j)}\). 
In practice, to ensure that each class's distributional difference with the rest is sufficiently optimized, we train for one specific class for 10 iterations before we turn to the next class. 

	%!TEX root = main.tex

\section{Experimental Setup}
\label{sec:exp}

In this section, we describe the datasets, baselines, and evaluation metrics used in our experiments.

\vpara{Datasets.}
%\paragraph{Datasets.}
We conduct experiments on four real-world datasets:
%, each of which is set in some classification scenario:
%To validate the effectiveness of our proposed framework, we construct experiments based on three real-world datasets, whose statistics are summarized in Table~\ref{tab:dataset}.  
\begin{itemize}[leftmargin=*]
	\item \textit{Adult~\citep{Adult:2013}.} This dataset 
	%is also known as the "Census Income" dataset. It 
	is extracted from the 1994 Census database.
	Each sample represents an individual and is classified based on whether the individual's annual income exceeds \$50,000. % based on census data.
	\item \textit{Statlog~\citep{Adult:2013}.} This is the German credit data. In this dataset, every individual is classified 
	as either good or bad in terms of credit risks. 
	%according to their credit risks. Each sample is labeled either good or bad in terms of credit risks. 
	%The dataset was provided with both categorical and numerical versions, in which the latter was used in our experiments. 
   \item \textit{Fraud.} This dataset is provided by PPDai, the largest unsecured micro-credit loan platform in China. 
   It consists of over 200,000 registered users and over 37 million call logs between them. 
   %We construct a communication network, where each vertex represents a user, and each edge denotes a call log from one user to another. 
   Each user's features are extracted from their basic information (e.g., age, gender, education, etc.) and call behavior  (e.g., call number, call duration, etc.).
   We aim to determine each user's credit risk by deducing whether they will default pn a loan for more than 90 days. 
   %Moreover, we have user labels according to their credit risk: users who ever defaulted a loan for equal to or more than 90 days are classified as ``fraudsters'', others as ``normal''.  
   %This is extracted from mobile network provided by PPDai~\footnote{The largest unsecured micro-credit loan platform in China.}. Each sample is a registered PPDai user. Attrbutes include basic information, e.g. age, gender, and education, and calling patterns extracted from the network. Each sample is classified according to their credit risk: users who ever defaulted a loan for equal to or more than 90 days are classified as "fraudster", and others as "normal". 
   \item \textit{Investor.} This dataset is provided by PPDai. It consists of almost 10,000 registered investors, and each of which is classified as investing over \$73,000. %(500K)%
   The attributes are the user's basic information (e.g., age, gender, residential place, house price) and behavior (e.g., frequency of use of the app, etc.)
\end{itemize}
%All datasets are set in some classification scenario. 
%For Adult, we are to classify high income individuals, while for PPDai and Statlog, we are to find individuals with higher credit risk. Positive instances in these datasets are respectively individuals with high income and high credit risk.
For all datasets, we choose Gender as the protected attribute. Female is the protected group in our tasks.
Note that although gender is a binary attribute in all the four datasets, we recognize that gender may not be binary.
%We call the target against which we desire to classify, i.e. income level and credit risk, as labels. 
%The attributes we desire to protect in learned representations are called protected attributes. 

\begin{table}[t]
\centering
\small
    \begin{tabular}{lrrrr}
        \toprule
        & Samples & Attributes & Protected (\%) & Positive (\%)\\\midrule
        Adult & 48,842 & 14  & 33.0 & 24.0 \\ % \hline
        Statlog & 1,000 & 20 & 15.0 & 30.0 \\ % \hline
        Fraud & 205,835 & 37  & 21.1 & 10.0 \\ % \hline
        Investor & 9,827 & 6 & 42.0 & 11.0 \\ \bottomrule

    \end{tabular}
 \normalsize
    \caption{Statistics concerning datasets. 
    The columns are the numbers of samples and attributes and percentages of protected group (female) and positive samples.
    \normalsize }
    \label{tab:dataset}
\end{table}

%\subsection{Experiment Setup}
\vpara{Tasks.}
%Our goal is to learn representations of individuals in each dataset, which maximizes efficiency of classification tasks against labels, while protecting sensitive information.  
%Therefore, 
In our experiments, we first learn the latent representations according to the given feature matrix $\mathbf{X}$, then we conduct the following classification tasks to validate the effectiveness of the learned representations:
\begin{itemize}[leftmargin=*]
\item \textit{Task I.} We use the representation methods to learn the latent features of the data and estimate the information loss by MSE and the distributional distance between different groups of the protected attribute.
\item \textit{Task II.} We further use the learned latent features to classify against the data label and examine the size of performance drop and whether the results meet fairness constraints. 
\end{itemize}

\vpara{Baseline methods.} 
In our experiments, we employ and compare the following different methods for representation learning.
\begin{itemize}[leftmargin=*]
	\item \textit{Original.} This method directly uses all features to train a classifier for task I. 
%	Then, it exclude the protected attribute and train another classifier for task II. 
	\item \textit{Original-P.} This process employs all features except for the protected one to train the classifier. %in Task I. 
	\item \textit{AutoEncoder.} This process uses an autoencoder to learn representations according to all features. 
	\item \textit{AutoEncoder-P.} This method uses an autoencoder to learn representations according to all features but the protected one.
	\item \textit{LFR.} This is a model for learning fair representations that was proposed by \cite{zemel2013learning}. In this method, instances are assigned to certain prototypes as latent representations. We use the parameters advised by the authors~\citep{zemel2013learning}.
	%This is composed of two mappings: an encoder which maps the input to a latent representation space; and a decoder carrying the inverse operation. Both mappings are realized using multi-layer perceptrons. 
   \item \textit{NRL.} This is the proposed fair representation learning method.
\end{itemize}

We also have compared NRL with the methods proposed by \citet{madras2018learning} and \citet{edwards2015censoring}. 
%, whose EMD distances are both over 0.2. 
However, we omit the detailed results considering their unstable performance. 
More specifically, these  two methods  
%We did not compare with other alternative baselines like \citet{madras2018learning} and \citet{edwards2015censoring} considering their unstable performance and 
%because in experiments we find them hard to tune and to achieve fairness they 
often require a more complex network architecture than ours. 
In addition, when using a three-layer neural network for autoencoder and the adversary on the Adult dataset, we found the adversary in \citet{madras2018learning} and \citet{edwards2015censoring}'s model indeed cannot predict the protected attribute. 
However, after the representation is learned, a classifier of the same architecture usually separates well the protected groups. 
%Specifically, the EMD distances with \citet{madras2018learning} and \citet{edwards2015censoring}'s methods are both over 0.2. We have given formulations why their methods are unstable in Section~\ref{sec:related}. 
%\reminder{why not baseline}

\begin{table}
	\centering
    \small
	\begin{tabular}{l|cccc}
		\toprule
		     & Adult & Statlog & Fraud & Investor \\ 
        \midrule
		AutoEncoder   & 0.02  &  0.02   & 0.02  &   0.04   \\ 
		AutoEncoder-P & 0.05  &  0.02   & 0.09  &   0.15   \\ 
        \midrule
		LFR  & 1.12  &  1.17   & 0.86  &   1.10   \\ 
        NRL  & 0.10  &  0.15   & 0.21  &   0.06   \\ 
        \bottomrule
	\end{tabular} 
	\caption{MSE loss when reconstructing the original features. 
	Lower values indicate less information loss of the corresponding representation of the original data. }
	\label{tab:metric:mse}
\end{table}

\begin{figure*}[t]
	\centering
	\subfigure[Statistical Parity]{
		\includegraphics[width=.22\linewidth]{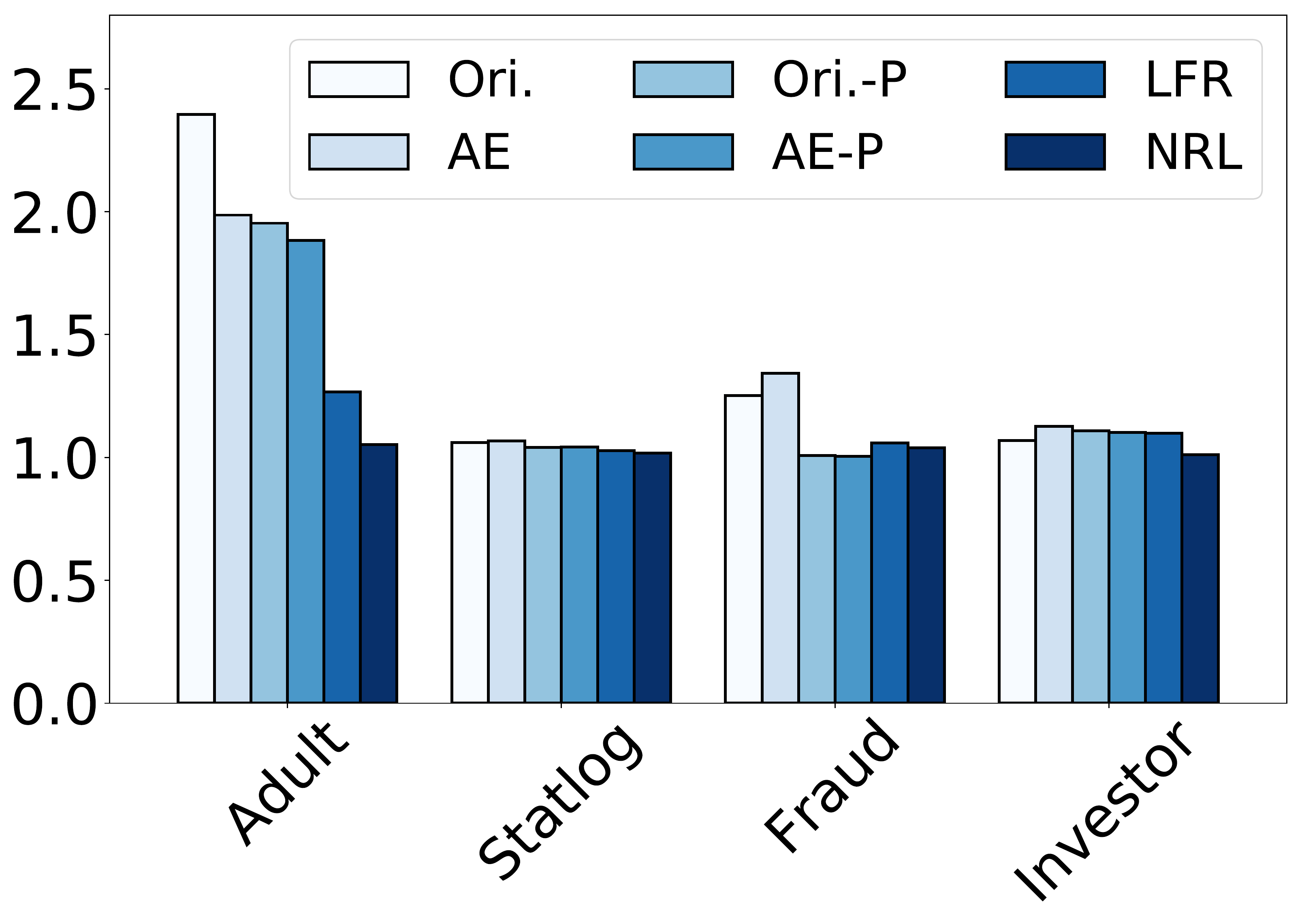}
		\label{fair:exp:fair:par}
	}
	\subfigure[EMD Distance]{
		\includegraphics[width=.22\linewidth]{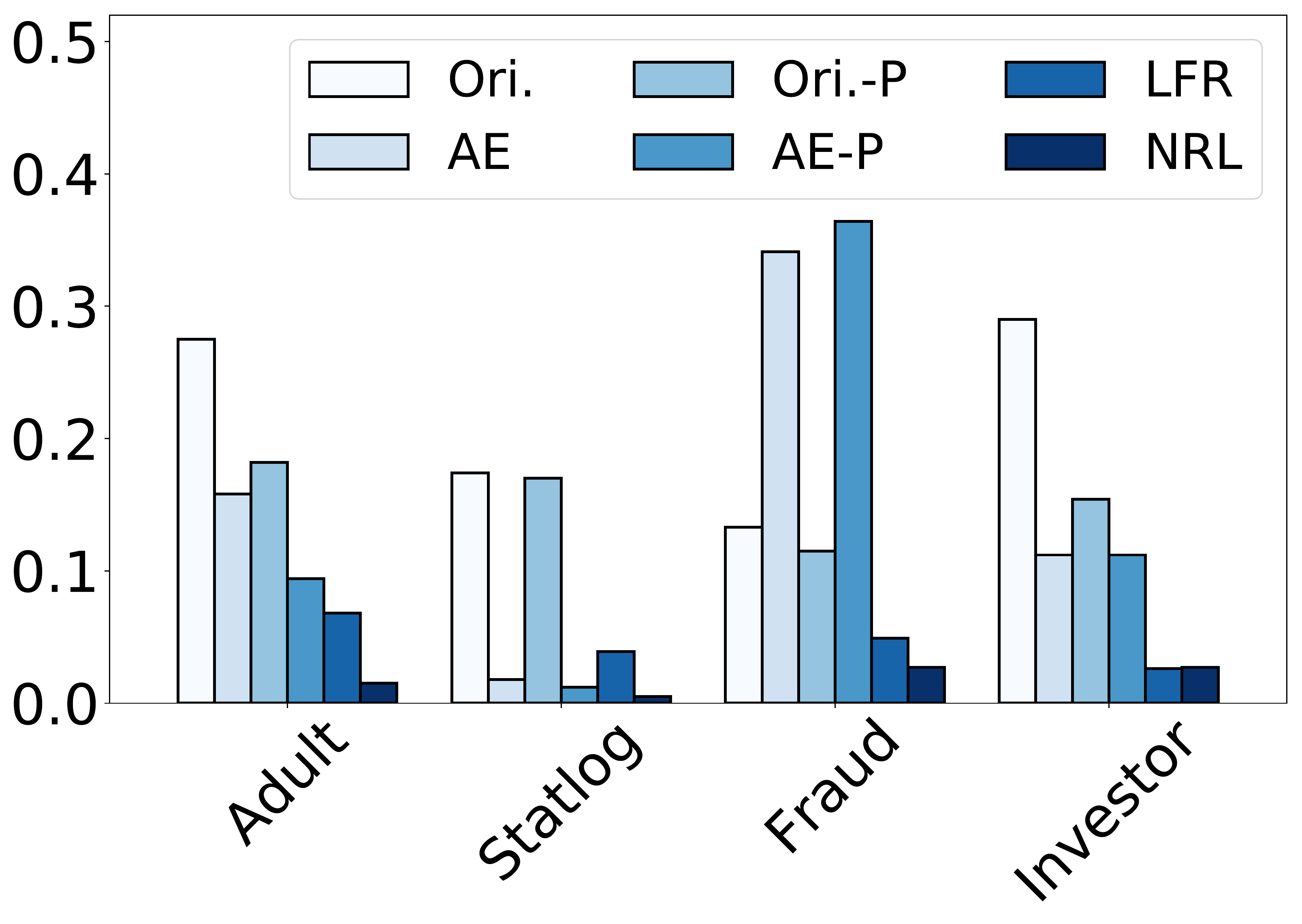}
		\label{fig:exp:fair:emd}
	}
	\subfigure[Consistency]{
		\includegraphics[width=.22\linewidth]{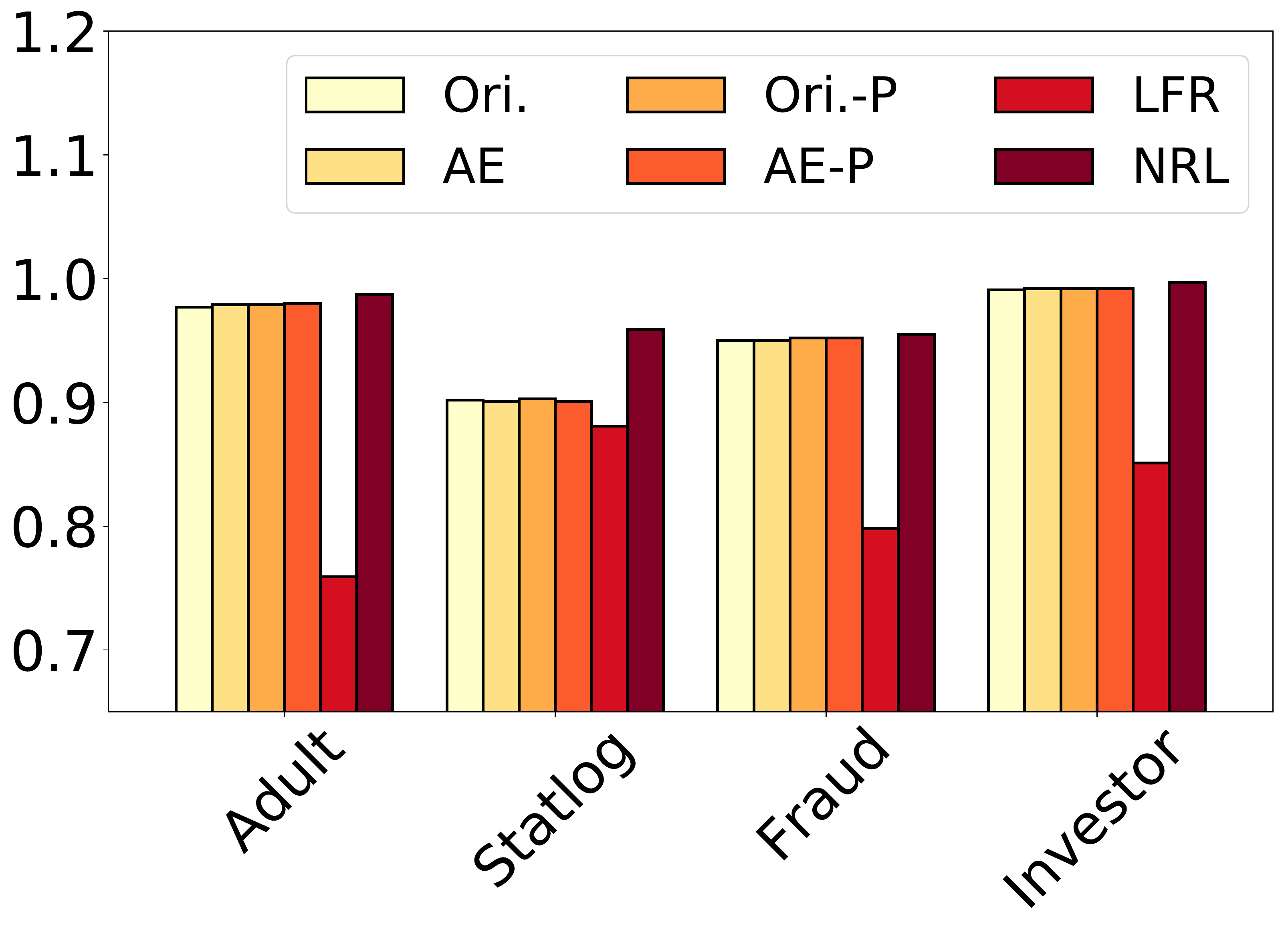}
		\label{fig:exp:fair:con}
	}
	\subfigure[Classification (F1-score)]{
		\includegraphics[width=.22\linewidth]{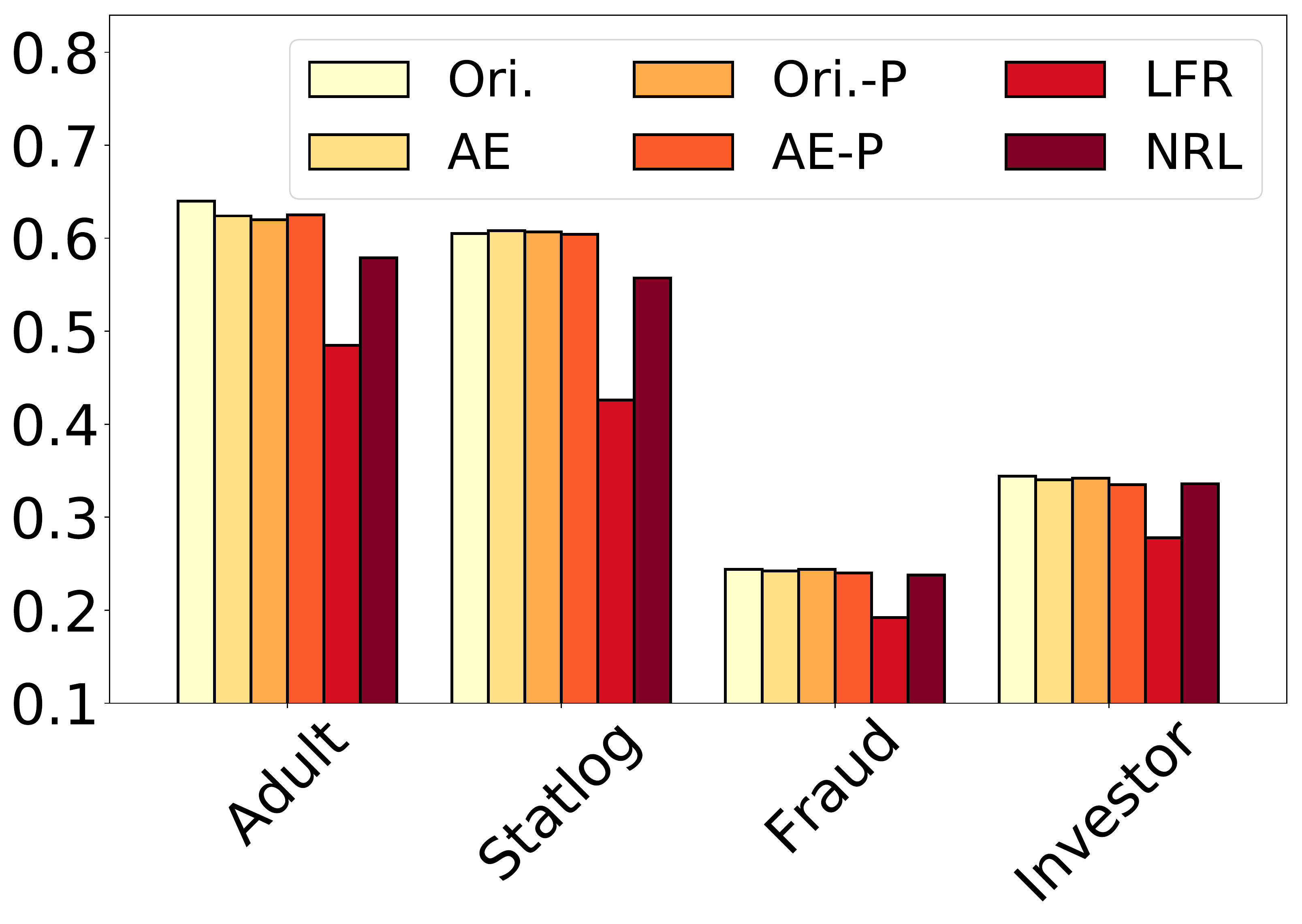}
		\label{fig:exp:fair:cls}
	}

\caption{Evaluation of fairness and classification performance. Ori., AE are short for Original and AutoEncoder. For statistical parity and EMD distance, lower values indicate fairer performance. For consistency, higher scores suggest better results in terms of individual fairness. For Classification, higher scores indicate better performance. }
\label{fig:exp:fair}
\end{figure*}

\hide{
\begin{table*}[t]
	\centering
	\begin{tabular}{l|cccccc}
		Method & Original & AutoEncoder & Original-P & Autoencoder-P & LFR & NRL \\ \hline
		Adult & 0.640 & 0.624 & 0.620 & 0.625 & 0.485 & 0.579 \\
		Statlog & 0.605 & 0.608 & 0.607 & 0.604 & 0.426 & 0.557 \\
		Fraud & 0.244 & 0.242 & 0.244 & 0.240 & 0.192 & 0.238 \\
		Investor & 0.344 & 0.340  & 0.342 & 0.335  & 0.278  & 0.336 

	\end{tabular}
		\caption{Performance on classification tasks. }
\label{tab:exp:classification}A
\end{table*}
}
\begin{figure*}[ht]
	\centering
	\subfigure[Adult]{
		\includegraphics[width=.22\linewidth]{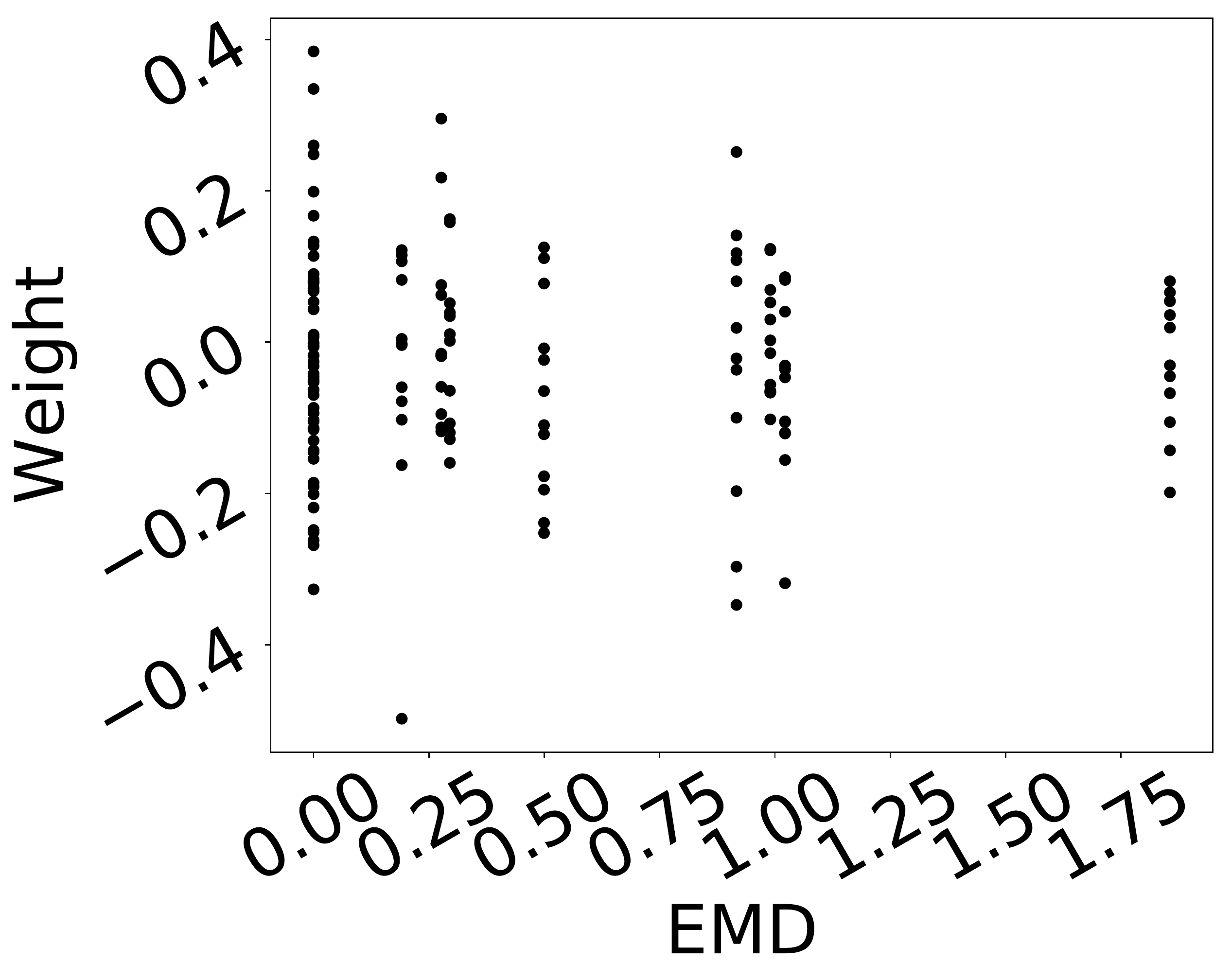}
	} 
	\subfigure[Statlog]{
		\includegraphics[width=.23\linewidth]{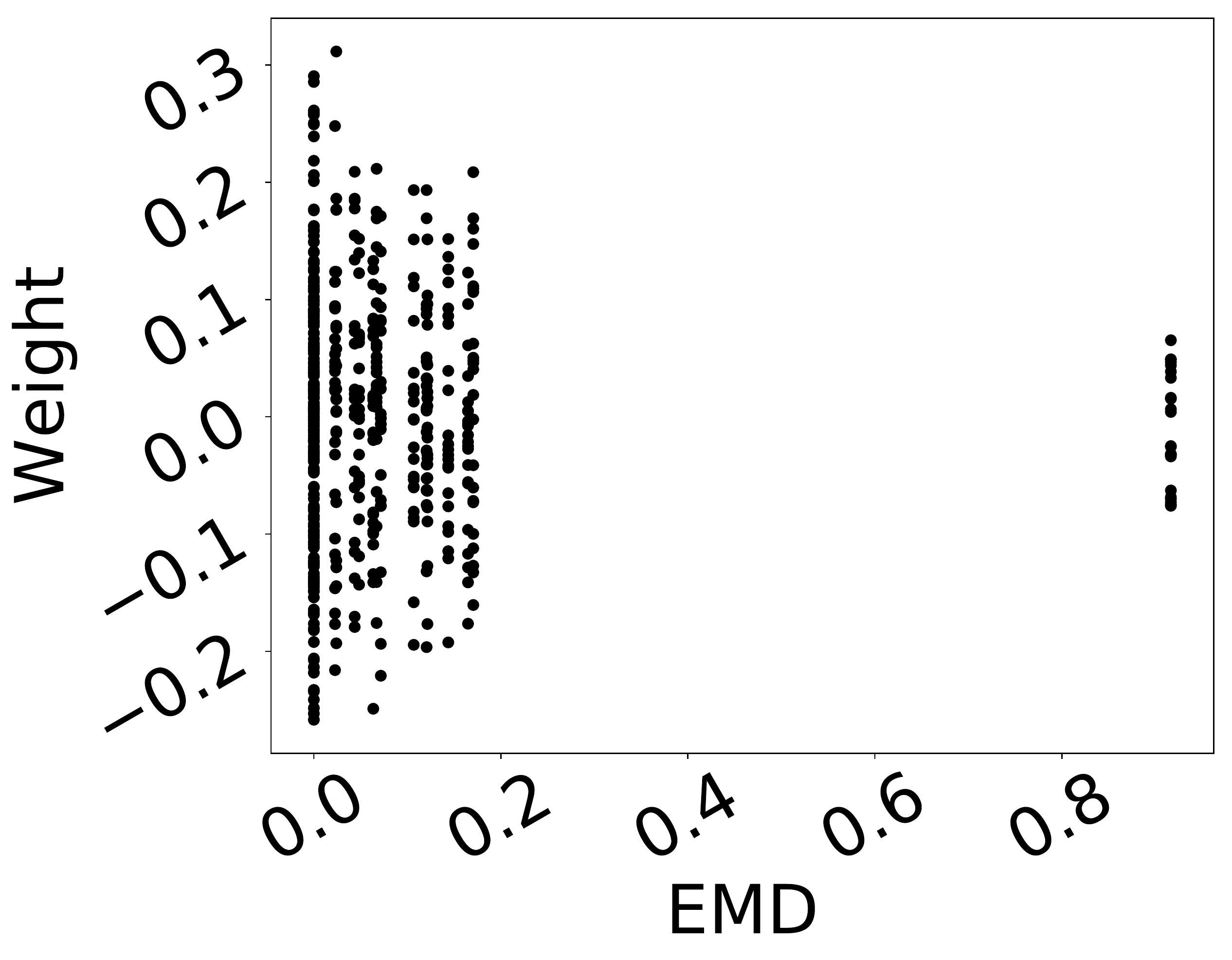}
	}
	\subfigure[Fraud]{
	\includegraphics[width=.23\linewidth]{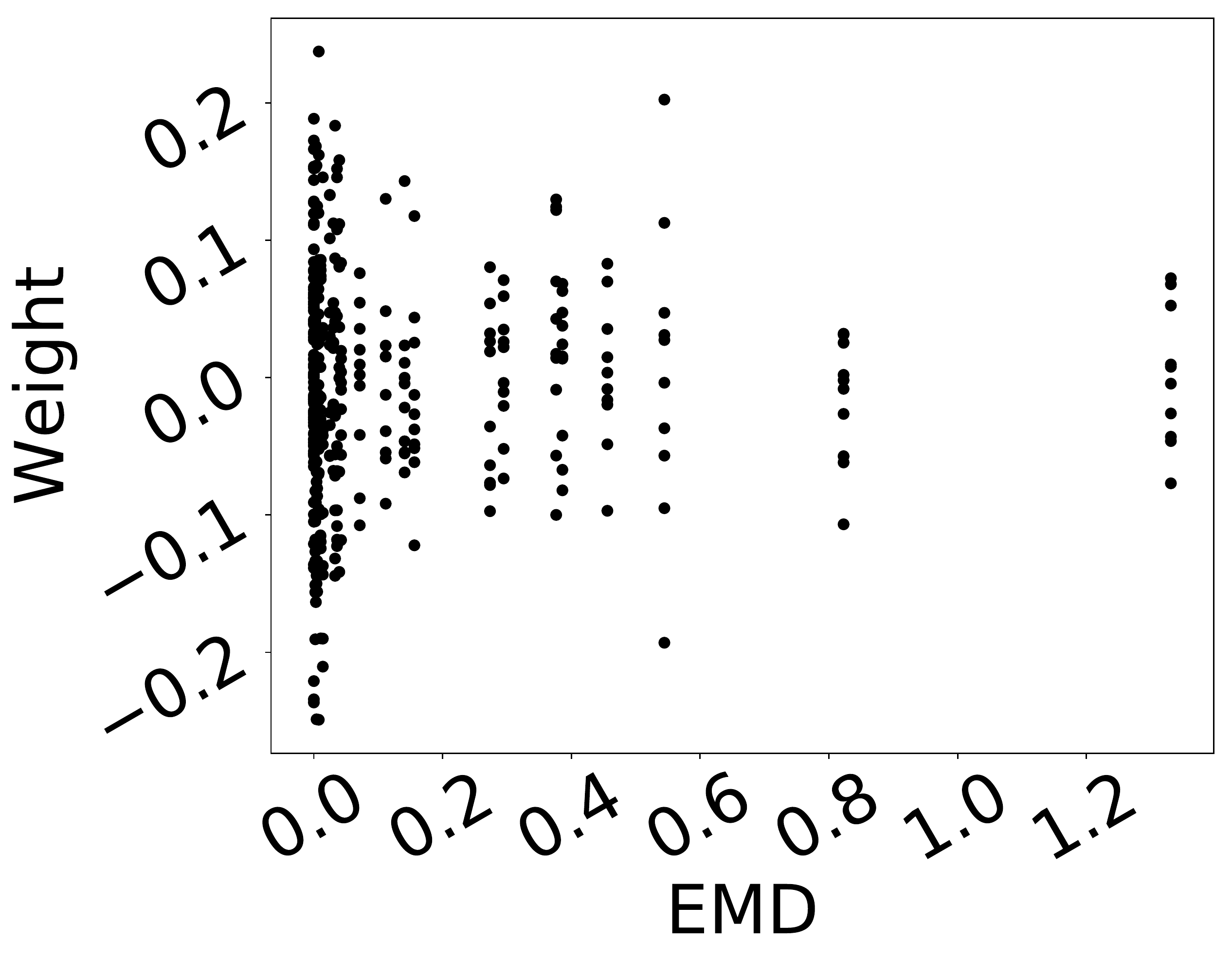}
	}
	\subfigure[Investor]{
		\includegraphics[width=.23\linewidth]{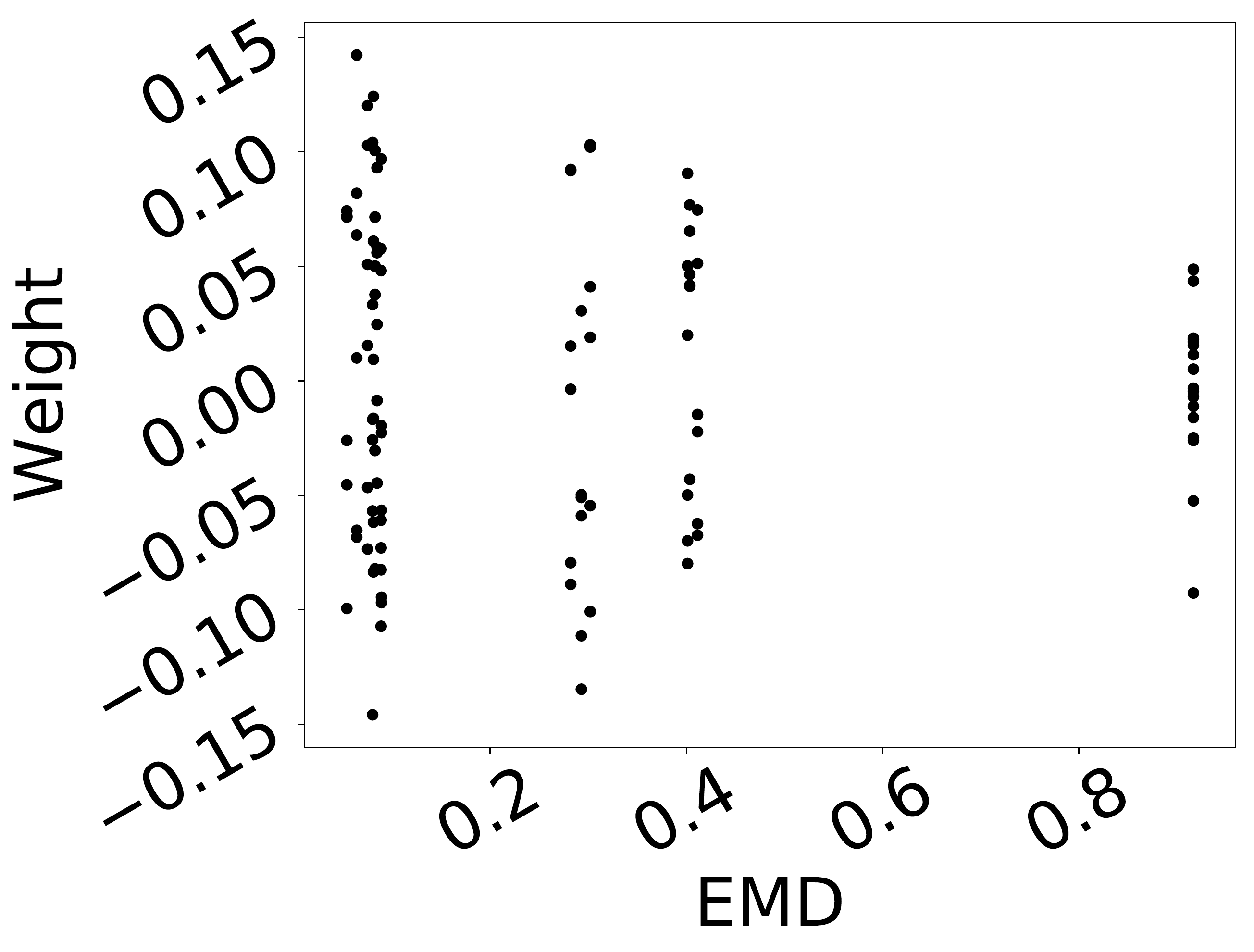}
	}
	\caption{EMD distance and the weight of the linear encoder. 
	% \small
     The x-axis is the EMD distance of each individual features between the protected groups. The y-axis is the weight corresponding to the feature in the linear encoders. The higher the absolute value, the more importance the model gives the feature in the representation. It can be observed that as the EMD increases, the weight converges to zero, indicating that the model is more reluctant in using that feature in the representation. \normalsize }
	\label{fig:illus:emd_weight}
\end{figure*}
\vpara{Methods of evaluation.}
\label{sec:exp:eval}
We evaluate the quality of the representation learning using the following metrics:
\begin{itemize}[leftmargin=*]
    \item The mean square error of the autoencoder, which estimates the information loss of the representation. 
    \item The EMD distance between the representations of different groups. 
\end{itemize}
In addition, in order to investigate the actual effect of our method, classification based on the learned latent representations is also studied. 
To obtain the classification results, 
%we employ LR(Logistic Regression), RLR(L2-regularized Logistic Regression), and CLR(Clamped Logistic Legression) on \(\mathbf{X}\) and \(\mathbf{U}\).
we employ RLR(\(\ell_2\)-regularized Logistic Regression, also known as the ridge logistic regression) on \(\mathbf{X}\) and \(\mathbf{Z}\). 
%The strength of the punishment is denoted by \(\C\) as the 
The following metrics are used to quantify the quality of classification.
\begin{itemize}[leftmargin=*]
	\item Classification performance against labels, evaluated by the F1-score. F1-score is calculated by the harmonic average of the precision and recall scores of the prediction.  
	
	\item The statistical parity score of the classification result against labels, defined as 
	\[
	\frac{\max{(\frac{1}{\nsamps_0}\sum_{p_i=0}\Psi(\mathbf{z}_i), \frac{1}{\nsamps_1}\sum_{p_i=1}\Psi(\mathbf{z}_i) )}}{\min{(\frac{1}{\nsamps_0}\sum_{p_i=0}\Psi(\mathbf{z}_i), \frac{1}{\nsamps_1}\sum_{p_i=1}\Psi(\mathbf{z}_i))}}
	\] 
	\item Consistency score of the classification, defined as 
	\[
	yNN = 1-\frac{1}{\nsamps}\sum_{i=1}^{\nsamps} \left\vert \hat{y}_i - \frac{1}{k}\sum_{j\in kNN(\mathbf{x}_i)} \hat{y}_j \right\vert 
	\]
	This equation evaluates the average differences in classification scores between a point and its \(k\)-nearest neighbors. A similar calculation is also used in \cite{zemel2013learning}. In our experiments, we use \(k=1\).
\end{itemize}

Specifically, to evaluate performance, fair representations are first learned and then fixed. On the learned latent features (or the original features, if no representation is learned), the EMD distance between the groups of the protected attribute and the MSE reconstruction loss are computed. 

Then, we train a logistic regression on the latent features to predict labels of data. The logistic regression predicts for each sample a score between 0 and 1 representing its probability of being positively labeled. Using the predicted score, the statistical parity score and the consistency score are calculated. 

\section{Experimental Results}
% \reminder{Would you take a look at this subsection in particular? RF}

We first compare our model performance with other approaches using a binary gender attribute, and then extend this to multi-class categorical variables.
We further show how our models can be used to discover bias in the dataset.

\subsection{Model Performance}
% Estimates of the metrics are given in Table~\ref{tab:metric:mse}, Table~\ref{tab:exp:classification}, and Figure~\ref{fig:exp:fair}, each being the average of 10 experiment outcomes. 
Since we attempt to impose fairness constraints, our method is necessarily outperformed in classification by methods without consideration of fairness, including Original, Original-P, AutoEnecoder, and Autoencoder-P. 
However, the advantage of our method is twofold:
1) it better satisfies fairness constraints than methods without consideration of fairness;
2) it preserves more information and better satisfies fairness constraints than prior approaches, namely, LFR.
% The goodness of our representation lies in two comparisons, i.e. it is fairer and its performance in classification and information preservation is as good as possible, or better than LFR in our case. 
 
\vpara{Reconstruction loss.}
Table~\ref{tab:metric:mse} shows the MSE loss of reconstructing the original features. As we expected, NRL sacrifices performance in information preservation at the cost of enforcing the distributions between two groups being identical, compared to AutoEncoder and AutoEncoder-P.
However, NRL preserves much more information than LFR. 

\vpara{Fairness constraints in classification.}
Figure~\ref{fig:exp:fair} shows the metrics to evaluate fairness, including the statistical parity score, the EMD distance, and the consistency score. For the former two metrics, lower values indicate better results: 
lower statistical parity means that the classification has no preference for one category of the protected attribute over the other;
lower EMD distance indicates that the information of the protected attribute is better obstructed. 
For consistency, higher values are better since they capture individual fairness of the classification: similar samples are treated similarly.

NRL consistently provides the best performance in all fairness constraints.
We note that AutoEncoder and  AutoEncoder-P are sometimes sufficient for achieving strong performance in statistical parity and consistency (Fraud and Investor), because both of these metrics rely on predictions and gender does not really matter in these prediction tasks.
In comparison, LFR is not even competitive in these easy cases.

Figure~\ref{fig:exp:fair:cls} shows the cost of fairness. Since we actively loses any information relevant to the protected attribute which is often correlated to the label, the performance in classification inevitability drops compared to methods without considering fairness. 
However, our method is better compared to LFR, indicating better information preservation, consistent to results in Table~\ref{tab:metric:mse}. 
Our results suggest that our method can serve as a benchmark to test whether simply approaches such as AE-P and Original-P are enough for achieving desired fairness constraints.

%However, our method  

\hide{

Table~\ref{table:metric:cls} further examines the realization and cost of fairness achieved by our model by exhibiting the classification results and the three metrics of fairness: statistical parity, consistency, and EMD. In Adult, Statlog, and Investor, our model greatly reduces the EMD and furthermore reduces statistical parity and increases consistency with regularized linear classifiers. 
Hence, in terms of fairness, it is a better method than applying the same classifier to the original data and plain autoencoders, which do not take fairness into consideration. 
In terms of classification performance, we are better in classification results with higher F1 and K-S scores compared with LFR. 
}
Our experiments evidence that in most cases, simply removing the protected attribute is insufficient. In three of the four datasets, AE-P and Original-P cannot be fairer than NRL. For the Investor dataset, AE-P obtains even lower F1-score and a higher MSE loss. 

The Fraud dataset deserves special attention: its statistical parity and consistency is already low enough due to the removal of the protected attribute. Representations learned by NRL and LFR could not be fairer than in the original dataset with the protected attribute removed. Although removing the attribute cannot keep the EMD low, if one is concerned only with classification results, such a simple modification is fair.

\vpara{How fair representations are generated.}
It is also of crucial interest that how the model chooses to represent the data. In this section, we give some concrete examples to illustrate how the model learns the proper representation of data that is fair. 

For each individual feature of a dataset, we explored the relationship between  EMD of that feature across the protected groups and the weight of that feature in the linear encoder. We scatter-plotted the EMD and the weights in Figure~\ref{fig:illus:emd_weight}. One may observe the trend that the linear encoder tends to assign smaller weights in terms of absolute value to features with higher EMD. Since features with higher EMD contribute more to the difference of the joint distribution of all features, it is an empirical evidence that the linear encoder attempts to suppress information related to the protected attribute with the help of the critic. 

\subsection{Multiclass Protected Attributes}
We take the race for an example of a multi-class protected attribute. In Adult dataset, the feature "race" may take as many as five categories: ``White'', ``Asian-Pac-Islander'', ``Amer-Indian-Eskimo'', ``Other'', and ``Black''.

%\reminder{The model setting, training procedure, and classification performance here is presented by text. Please check if it is clear and unambiguous. } 
For model implementation and parameters in this experiment, the encoder is a three-layer neural network with ReLU nonlinearity. The dimensions of all the hidden layers and the output layer are 10. The decoder is the reversal of the encoder. Finally, the critic is a linear mapping to \(\RR\), as in previous univariate experiments, and the weight of the loss, \(\alpha\), is set as 1000. 

For evaluation, the metric which measures the statistical fairness of a prediction, \emph{statistical parity}, can be thus defined in the multi-class case: 
\beq{
	\frac{\max_{1\leq i\leq0 n_{p}-1} (\frac{1}{n_j} \sum_{p_i=j} \Phi(\mathbf{z}_i)) }
	{\min_{1\leq i\leq n_{p}-1} (\frac{1}{n_j} \sum_{p_i=j} \Phi(\mathbf{z}_i))}
}
\noindent which is the ratio of the maximum to the minimum of the average predicted scores of each class derived by the protected attribute. 

Originally, the F1-score of the classification task is as high as 0.64, as indicated in Figure~\ref{fig:exp:fair:cls} Through our learned representation, the F1-score is dropped to 0.42 at the cost of fairness. This is lower than shown in Figure~\ref{fig:exp:fair:cls}, indicating that race is more costly than gender when concealed. 
As a comparison, we also trained vanilla autoencoder on the Adult dataset, with the same architectures for the encoder and the decoder. However, it is hard to converge to a more efficient representation than a simple linear autoencoder, and the average of our attempts yields 1.9 MSE and 0.530 F1 score. Therefore, we also trained a linear autoencoder for comparison. 

Experiment results are listed in Table~\ref{tab:exp:multiclass}, in which we listed two metrics: average prediction score, obtained by a regularized linear classifier based on specific representations; and one-vs-rest EMD, obtained by calculating the distribution difference between one specific class and all others. Using the average prediction score, we may compute the statistical parity for the four methods: 1.70 for original features, 1.77 for multilayer autoencoder, 2.18 for linear autoencoder, and 1.01 for NRL. 
This indicates that NRL may efficiently prevent discrimination when the protected attribute is multiclass.  

One should note that when race information is not hindered by NRL, all other three methods tend to assign lower probabilities of high income to groups of American-Indian-Eskimos and Blacks. This reflects that such discriminatory tendency exists in the data. Our model shows an alternative reality: what would happen if all except race is considered?
In Table~\ref{tab:exp:multiclass}, we see that the probabilities across ethnic groups are leveled around 0.24, a rather high probability, and is close to the White and Asian-Pacific-Islanders, as predicted by original features. 
 The results may shed light on how the economic status of some individuals or ethnic groups as a whole would have been were they not discriminated based on their race. 

\begin{table}[tbp]
	%\resizebox{0.98\linewidth}{!}{
	\begin{tabular}{|l|l|c|c|c|c|c|c|}
		\hline
		Metric                     & Method &  AIE  &  API  & Black & Other & White  \\ \hline
		\multirow{4}*{ Avg. Score} & Ori.   & 0.185 & 0.261 & 0.153 & 0.160 & 0.255  \\ \cline{2-7}
		                           & MAE    & 0.164 & 0.291 & 0.174 & 0.178 & 0.249  \\ \cline{2-7}
		                           & LAE    & 0.359 & 0.544 & 0.280 & 0.613 & 0.314 \\ \cline{2-7}
		                           & NRL    & 0.239 & 0.240 & 0.238 & 0.237 & 0.239 \\ \cline{1-7}
		\multirow{4}*{ 1vsR EMD }  & Ori.   & 0.468 & 0.430 & 0.281 & 0.418 & 0.274 \\ \cline{2-7}
		                           & MAE    & 0.097 & 0.098 & 0.096 & 0.151 & 0.075 \\ \cline{2-7}
		                           & LAE    & 0.164 & 0.291 & 0.174 & 0.178 & 0.249 \\ \cline{2-7}
		                           & NRL    & 0.001 & 0.001 & 0.002 & 0.002 & 0.0017  \\ \cline{1-7}
	\end{tabular}
	%}
	\caption{Results for multiclass protected attribute. 
	% \small 
    We listed two metrics for fairness: Avg. Score, which stands for average prediction score, and 1vsR EMD, which stands for one-vs-rest EMD distance. AIE stands for Asian-Indian-Eskimo, and API stands for Asian-Pac-Islander. Four methods are considered: original features, MAE (multilayer autoencoder), LAE (linear autoencoder), and our approach. 
    \normalsize  }
	\label{tab:exp:multiclass}
\end{table}

\subsection{Discovering Biases}
%\reminder{Is the presentation clear, reasonable, and persuasive? Especially in ``locating discriminated individuals''. }
% \vpara{Augmenting discrimination by flipping labels.}

Another important problem in the field of algorithmic bias is to discover potential biases.
%  discrimination discovery. 
We have so far presented some concrete examples of how the model can ``correct'' the labels of some females whose labels are ``unfairly'' negative, while these females have similar male counterparts with positive labels. 
To see if our model can be used for discovering biases, 
we explicitly ``discriminate'' some females by flipping their labels.
%  of the females. 
Specifically, we match each male to its nearest female in terms of Euclidean distance and select the pairs in which both the male and female are positively labeled. 
We use two approaches to examine the robustness of our framework to such added discrimination: 1) whether we can predict the real label;
2) whether we can detect such discrimination.

% We flip the labels of the female, and see if based on our representation, we can obtain positive predictions for these females. 

\vpara{Detecting flipped labels.}
We apply a linear classifier to both our learned representation and the original features. The regularization strength for each classifier is tuned to maximize the proportion of which the flipped labels are corrected. For Adult dataset, 1399 such pairs are selected, and the same amount of female's labels are manually changed to be negative. %to 0. 
Based on our learned representation, 84.4\% of the flipped females are predicted as positive by a linear classifier, while on the original features, % the figure is
only 40.6\% are. Similarly, 93 pairs are selected for Statlog dataset, and learning on our representation corrects 46.2\% of the flipped labels, while on the original features, only 8.6\% are successfully classified as positive. 

\vpara{Locating discriminated individuals. } The problem of locating discriminated individuals may also be conveniently formulated as locating ``mislabeled'' samples due to the influence of the protected attribute that ought to have no influence. 
In practice, an important problem is how to efficiently locate samples that are probably mislabeled, and human experts may proceed to check the labels in detail. 
In our experiment setting, it is desirable to locate the females whose labels we manually flipped. 

An individual is likely to be discriminated or mislabeled if when the protected attribute is not considered, he or she has a high prediction score compared to when the protected attribute is considered. 
Suppose the prediction score based on original features is \(p_o\) and the score based on our fair representation is \(p_f\). We define the following statistics to quantify the ``strength'' of discrimination against an individual as follows: 

\beq{
s_d = 1-\frac{p_o}{p_f}
}

\begin{figure}[t]
	\centering
	\subfigure[Adult]{
		\includegraphics[width=.45\linewidth]{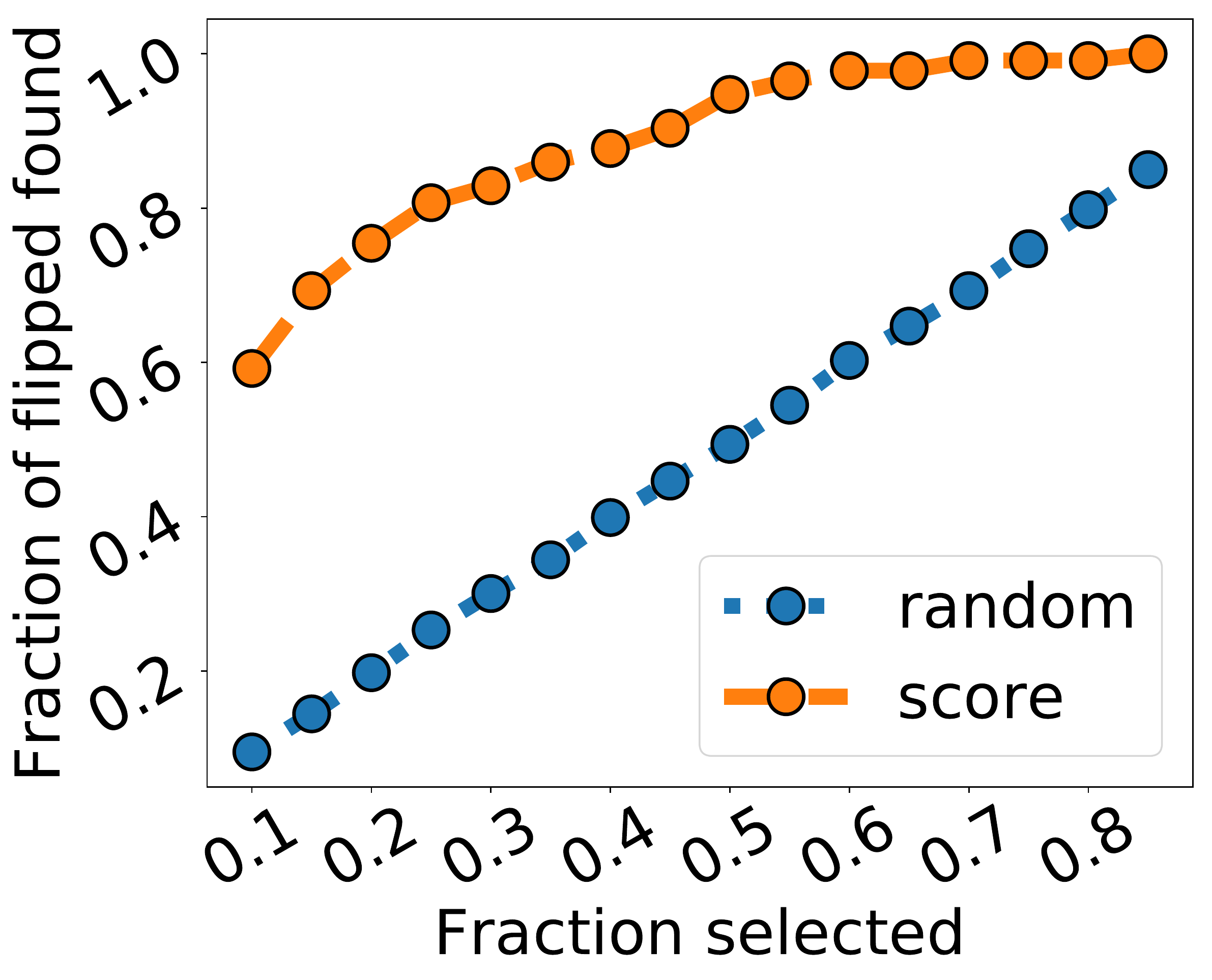}
	}
	\subfigure[Statlog]{
		\includegraphics[width=.45\linewidth]{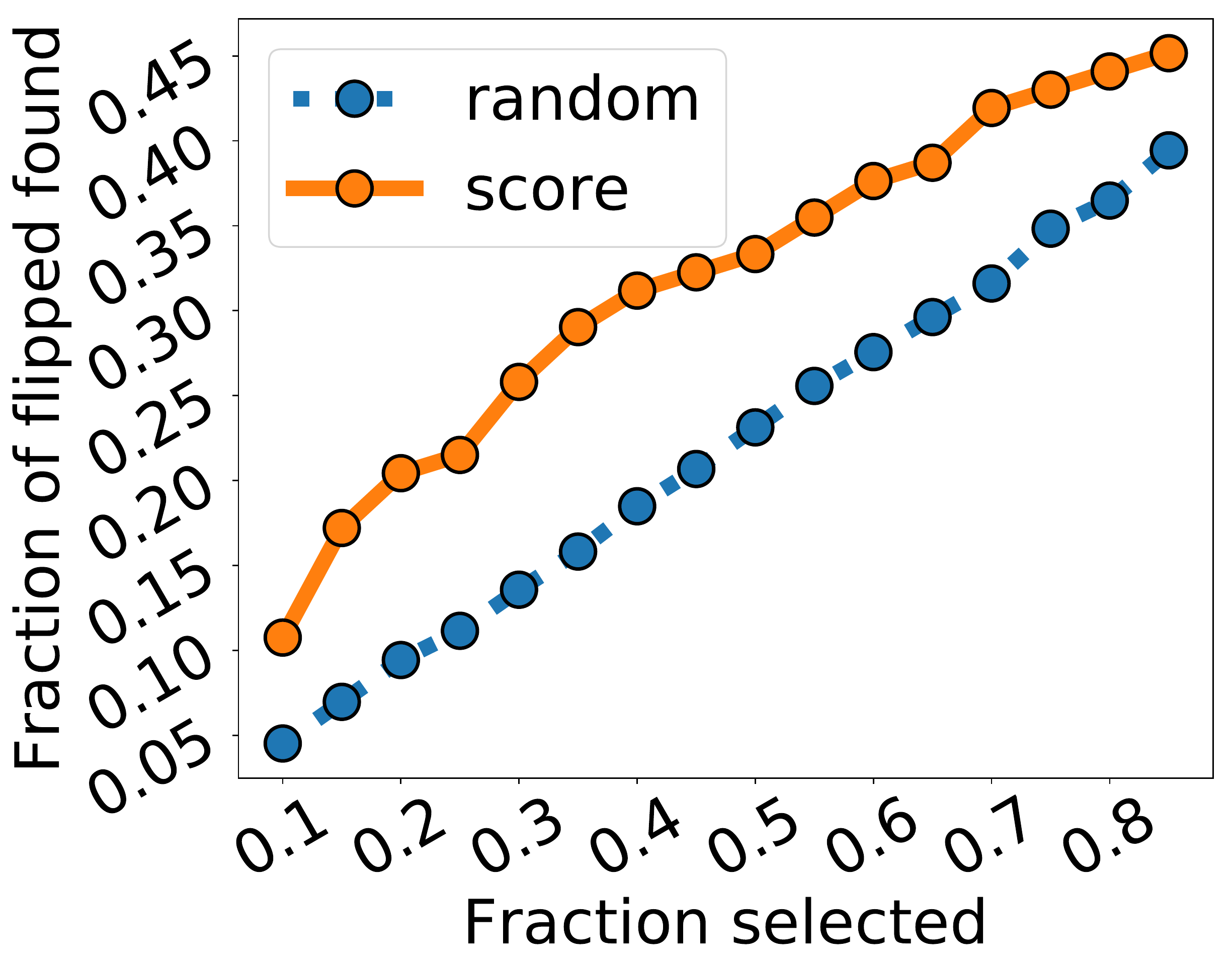}
	}
	\caption{Locating discriminated individuals. The x-axis is the fraction of females selected, and the y-axis is the fraction of ``discriminated'' (flipped) females discovered. \normalsize }
	\label{fig:locate_disc}
\end{figure}

Then we sort the female individuals by \(s_d\) in descending order, and check the proportion of flipped females among females with the highest $s_d$. The results in Adult and Stalog are in Figure~\ref{fig:locate_disc}.	
Individuals identified with high $s_d$ are clearly much more likely to be these ``discriminated'' ones, compared to a random baseline.
% Uniformly random sampling of the female individuals is compared to our approach. 
% The result 
% %Figure~\ref{fig:locate_disc} 
% indicates that our approach may also be applied to locate discriminated individuals more efficiently. 

	%!TEX root = main.tex

\section{Conclusion}
\label{sec:conclusion}

In this paper, we proposed learning a latent representation of attributes that achieves fairness by obstructing information concerning a protected attribute as much as possible, while preserving other useful information.
Our method pre-processes the data and removes both direct and indirect discrimination for downstream tasks.
%  such that no discriminatory result may be obtained by downstream tasks. 
% The problem is formulated as an minimax optimization problem
We formulate this problem in a minimax adversarial framework
 by learning a transformation of features such that the distributions between different groups of the protected attribute, e.g., male and female, are indistinguishable. 
%The Wasserstein Distance, which is theoretically linked with two fairness constraints, statistical parity and individual fairness, is minimized in order to achieve this goal. 
%The Wasserstein Distance is theoretically linked with two fairness constraints, statistical parity and individual fairness. It can be shown that by the Kantorovich-Rubinstein Duality~\citep{villani2008optimal}, reducing the Wasserstein Distance in the represenations, in combination with a regularized classifier for prediction, one can achieve both individual fairness and statistical parity in the classification results. We performed various experiments to verify this. 
This framework provides theoretical guarantee on statistical parity and individual fairness, and also achieves strong empirical results on four real-world datasets.
We also note that for some tasks where protected attributes do not play any role, it seems sufficient to remove the protected attributes.
Our method can be used as a benchmark to detect these cases.

For future work, it is interesting to explore if our approach 
%of learning fair representations may 
can be extended to applications beyond fairness. 
For example, it is often useful to infer the counterfactual scenario: what would happen when one variable is absence? With that variable as the protected attribute, one may further analyze the problem. 
Another promising direction is to further consider continuous protected attributes or discrete ones with a large value range, such as age. 
%While learning representations in a latent space is useful for ensuring fairness in consequential data mining tasks, it is sometimes more desirable to simply modify the data in the original feature space such that the Wasserstein distance is minimized. Furthermore, our methods are concerned with only categorical protected attributes. A promising improvement of our model would be the extension to continuous protected attributes or discrete ones with a large value range, such as the age. 
%We would leave these improvements as our future work.
		\bibliographystyle{ACM-Reference-Format}
	{
	\balance
	\bibliography{ref}
	}
	\newpage
	%!TEX root = main.tex

% \onecolumn
\appendix
\section{Supplementary Materials}
\label{sec:supp}

\subsection{Model Parameters.} 
We employ a simple linear layer for the encoder, the decoder, and the critic. Thus, the representation \(\mathbf{U}\) is simply a linear transformation of the input \(\mathbf{X}\). 

The continuous variables of all four datasets are scaled by their mean and standard variance. For NRL and AE, the dimension for the feature space is 12 for Adult, 23 for Statlog, 30 for Fraud, and 5 for Investor. The dimension for AE-P is reduced by one. \(\alpha\) for NRL is 10 for Adult, Fraud, and Investor, and 100 for Statlog. The inverse of the regularization strength for logistic regression for all datasets is 0.01. 
\subsection{Details of Evaluation.}
We used NRL to learn fair representations on the entire dataset. After that, two classifiers are trained on the learned representations to evaluate performance as described in Section~\ref{sec:exp:eval}. 

For each classifier, the parameters are trained on 70\% of the data, i.e. the training set, randomly and uniformly sampled from the whole dataset. The scores are calculated on the remaining test set. This process is repeated 5 times and the scores are averaged. 

The whole process from training fair representations to training classifiers is repeated 10 times and the final displayed scores in Table~\ref{tab:metric:mse} and Table~\ref{table:metric:cls} is the average of the 10 repeated experiments.

\subsection{Parameter Analysis}
\begin{figure*}[h]
	\centering
	\subfigure[Statlog]{
		\includegraphics[width=.22\linewidth]{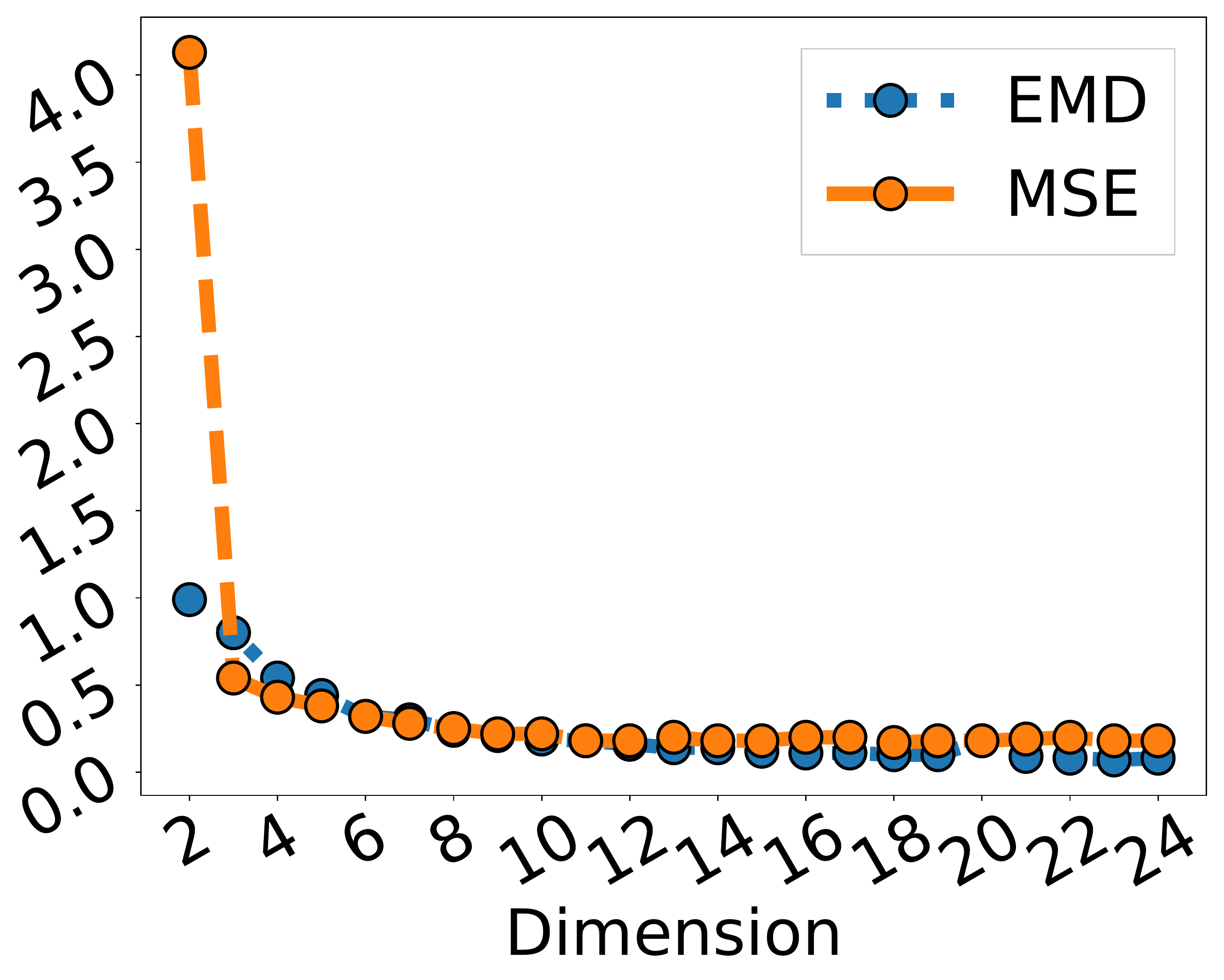}
	}
	\subfigure[Adult]{
		\includegraphics[width=.22\linewidth]{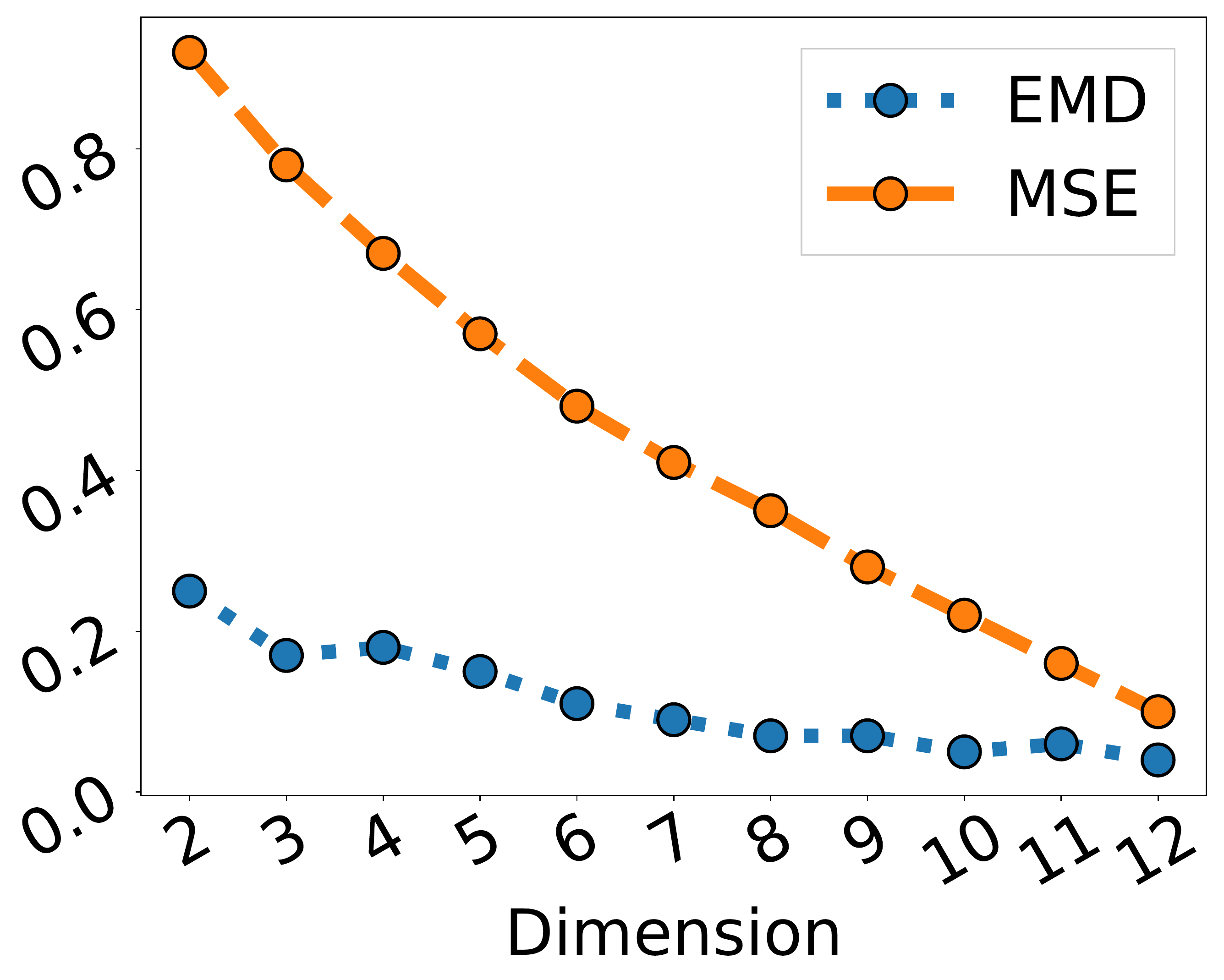}
	}
	\subfigure[Investor]{
		\includegraphics[width=.22\linewidth]{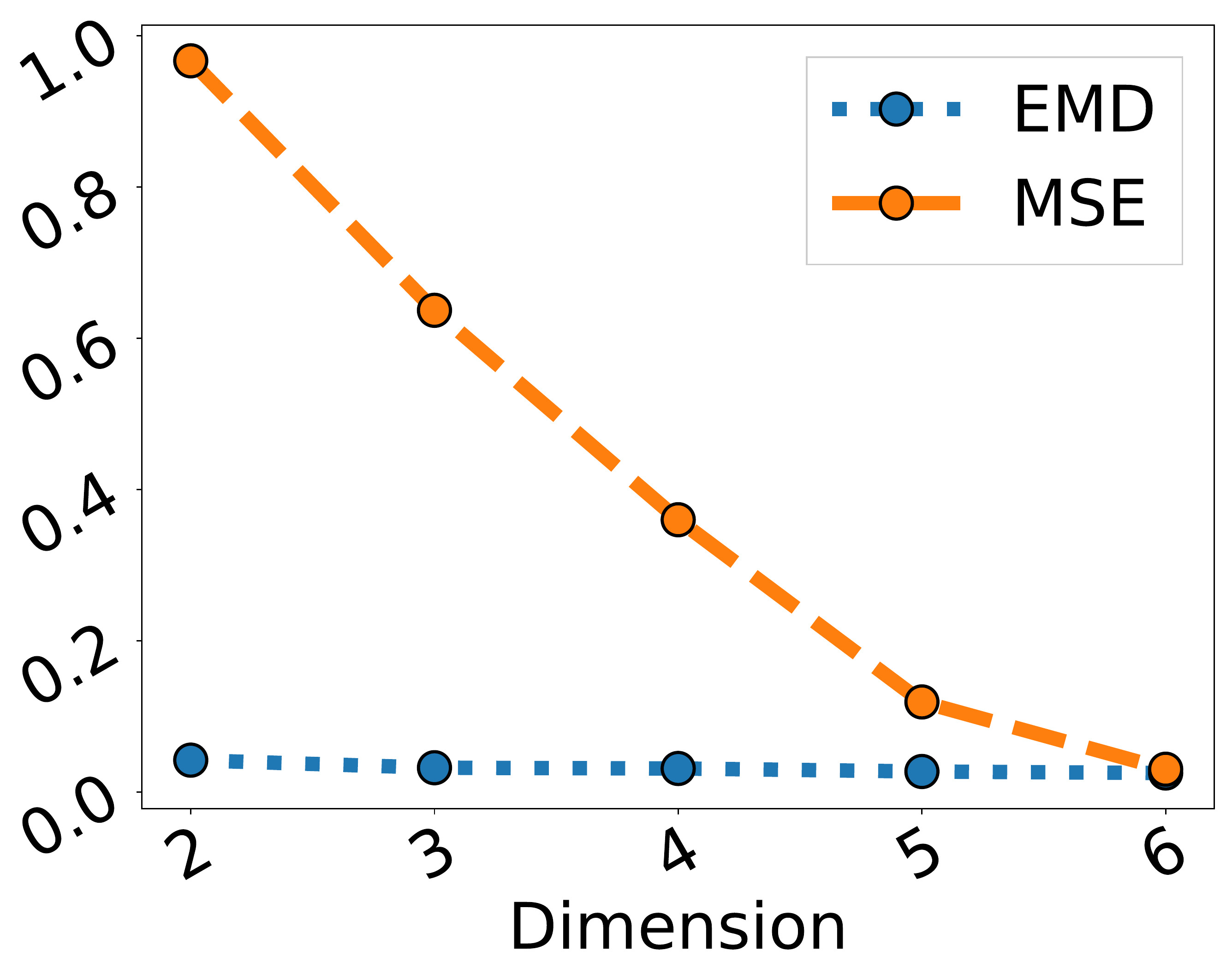}
	}
	\subfigure[Fraud]{
		\includegraphics[width=.22\linewidth]{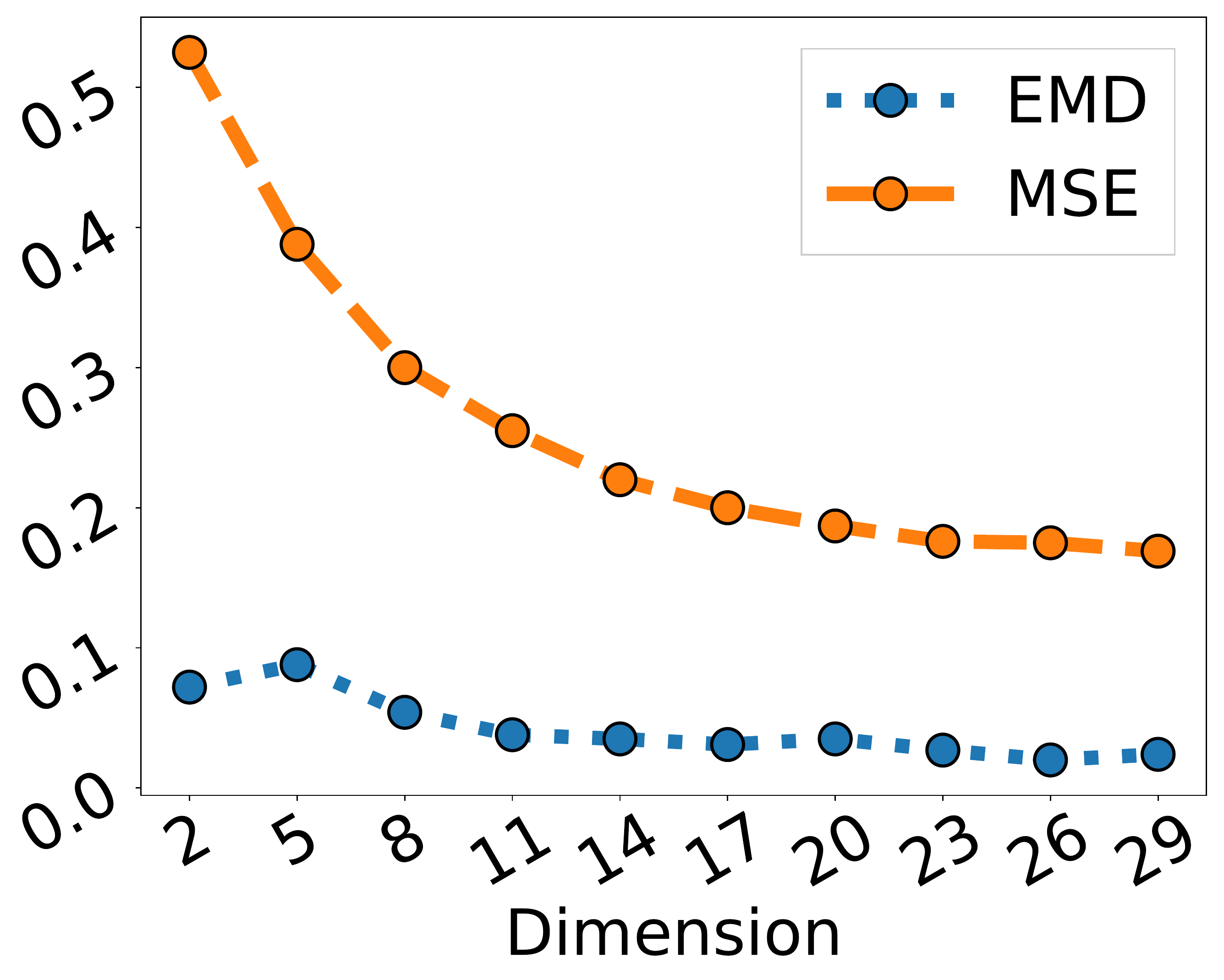}
	}
	\caption{Dimension and MSE and EMD. \small Generally, as the dimension increases, both MSE and EMD reduces. \normalsize  }
	\label{fig:params:dim}
\end{figure*}

\begin{figure*}[ht]
	\centering
	\subfigure[Statlog]{
		\includegraphics[width=.2\linewidth]{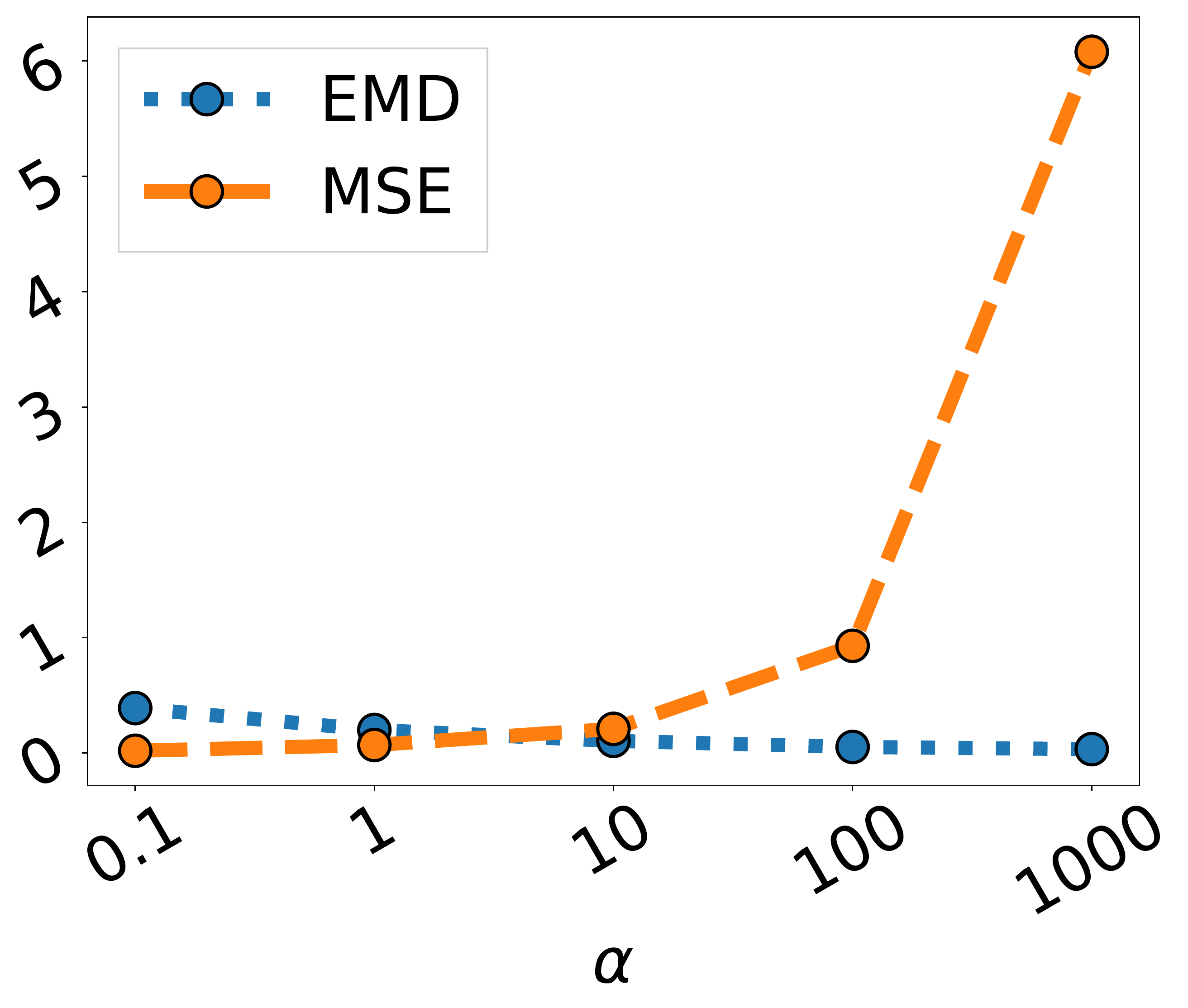}
	}
	\subfigure[Adult]{
		\includegraphics[width=.22\linewidth]{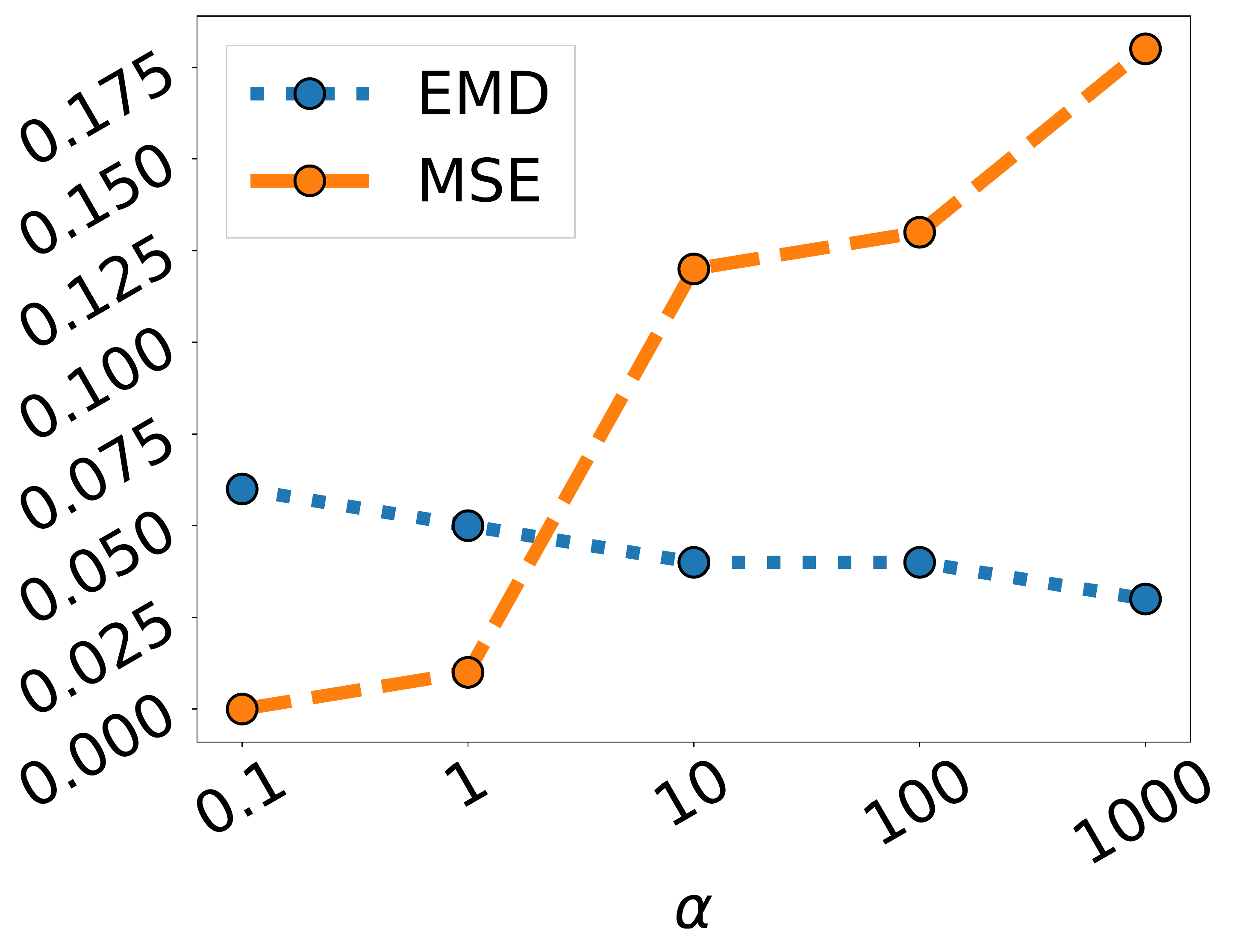}
	}
	\subfigure[Investor]{
		\includegraphics[width=.22\linewidth]{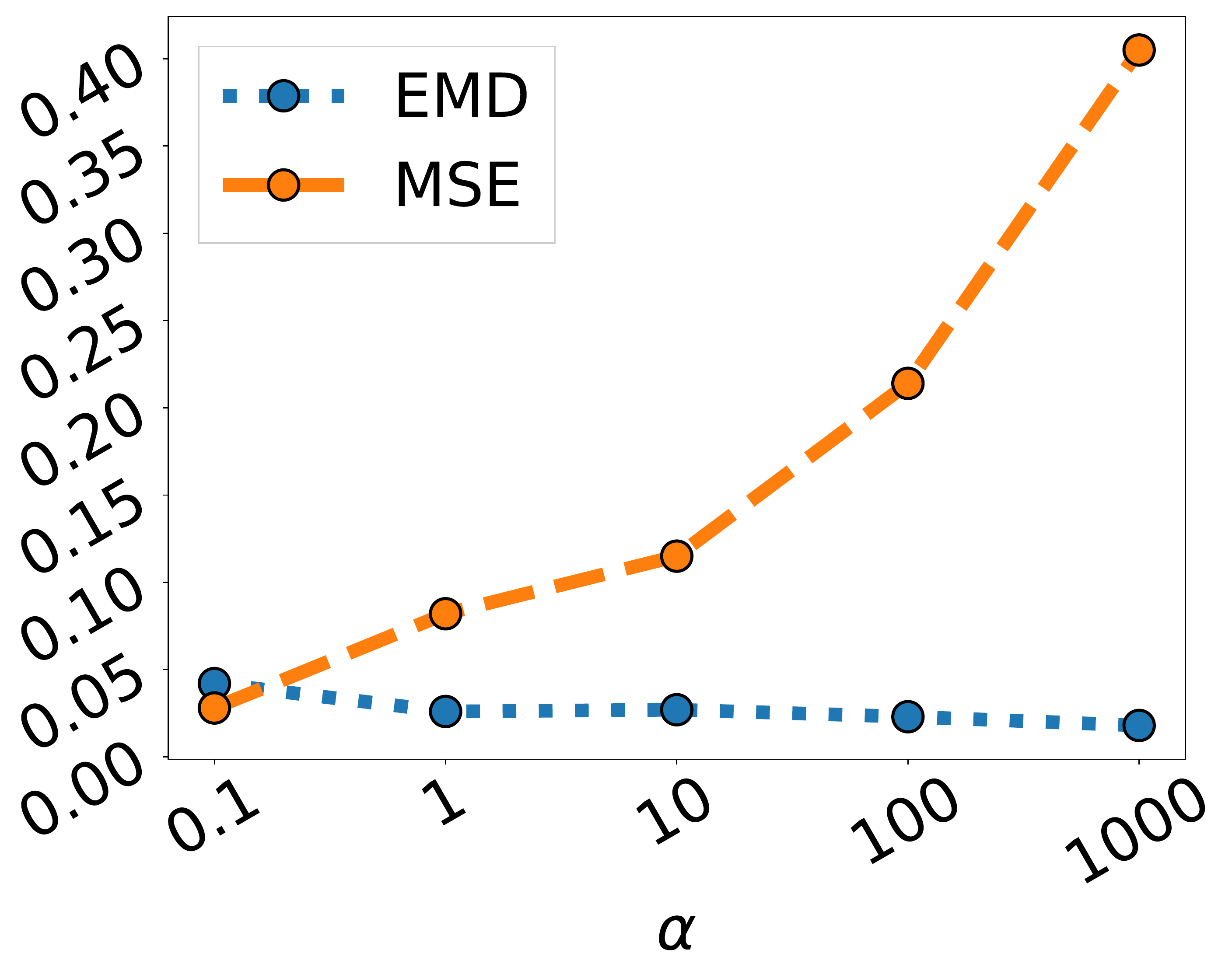}
	}
	\subfigure[Fraud]{
		\includegraphics[width=.22\linewidth]{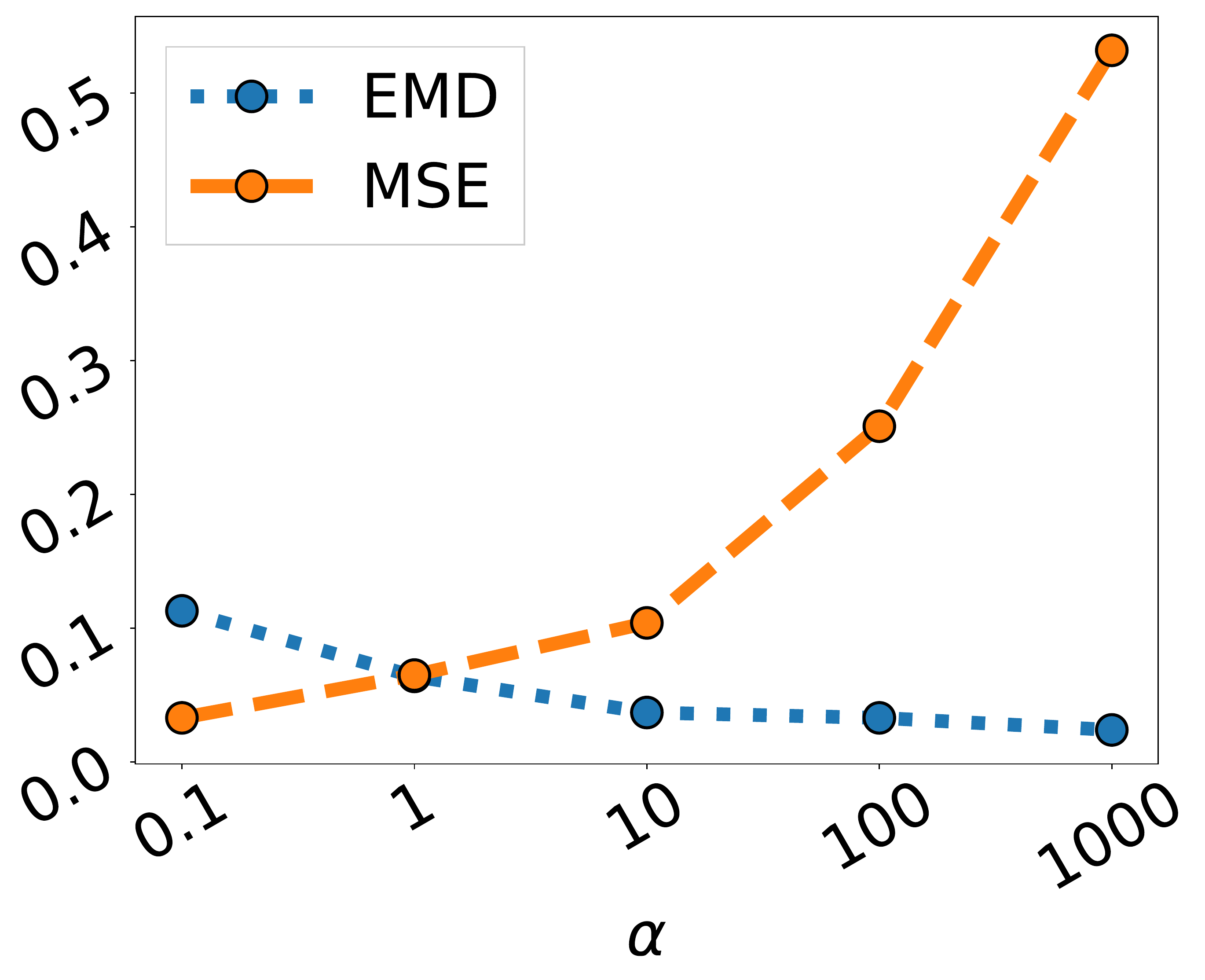}
	}
	\caption{Parameter \(\alpha\) and MSE and EMD. \(\alpha\) controls the strength of fairness constraints; higher \(\alpha\) enforces the model to generate fairer representation by reducing EMD.}
	\label{fig:params:alp}
\end{figure*}

\begin{figure*}[t]
	\centering
	\subfigure[Statlog]{
		\includegraphics[width=.22\linewidth]{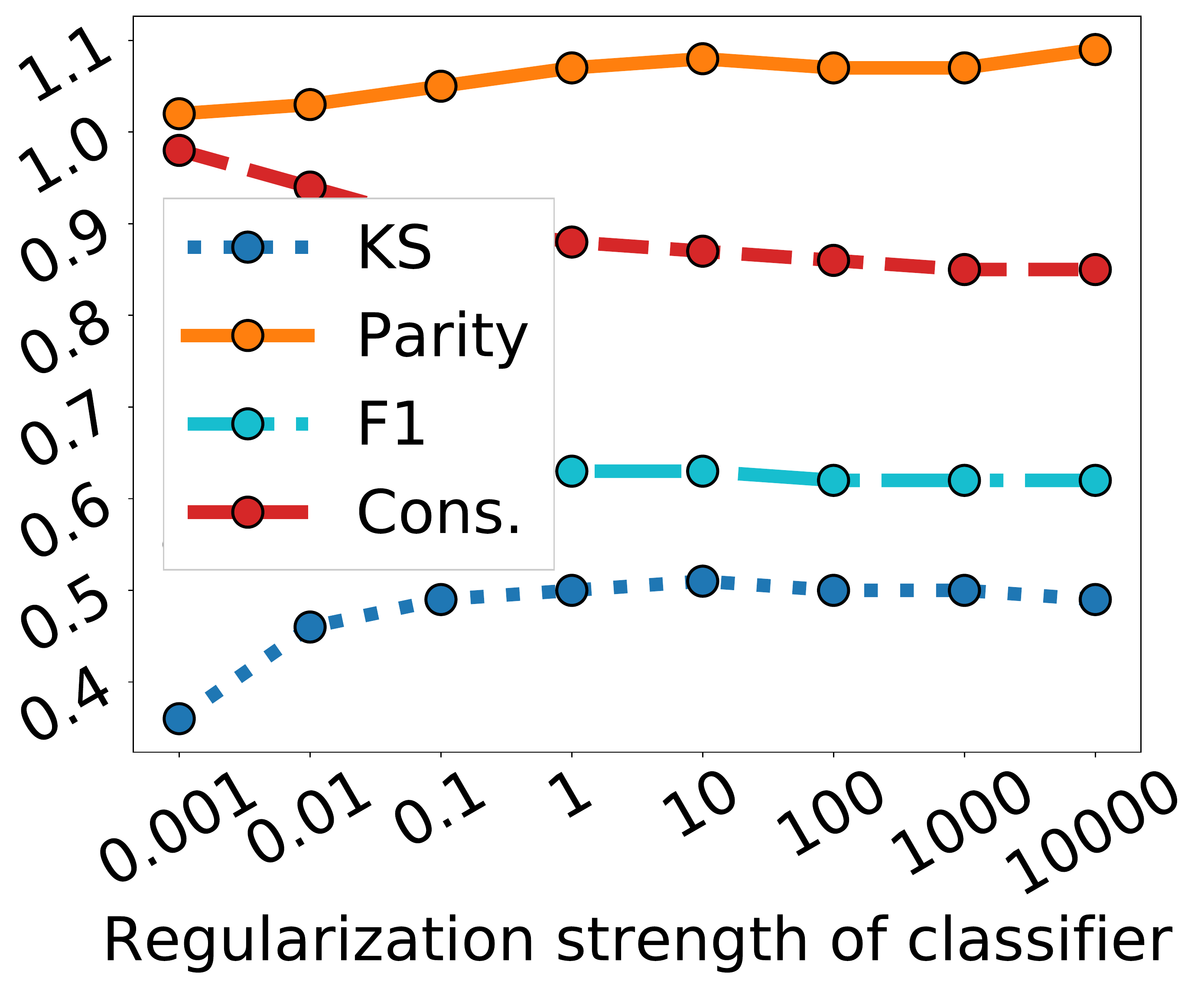}
	}
	\subfigure[Adult]{
		\includegraphics[width=.22\linewidth]{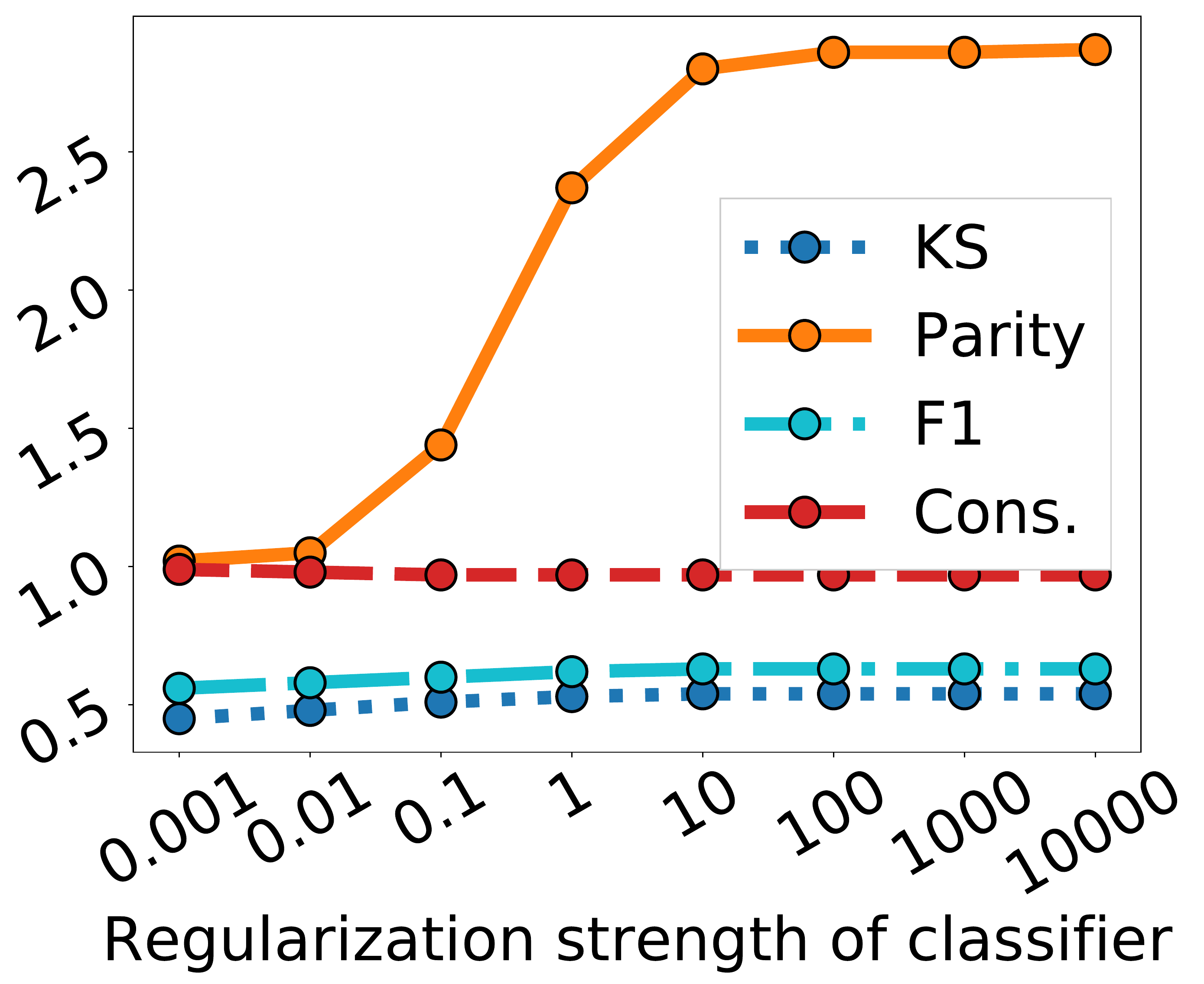}
	}
	\subfigure[Investor]{
		\includegraphics[width=.22\linewidth]{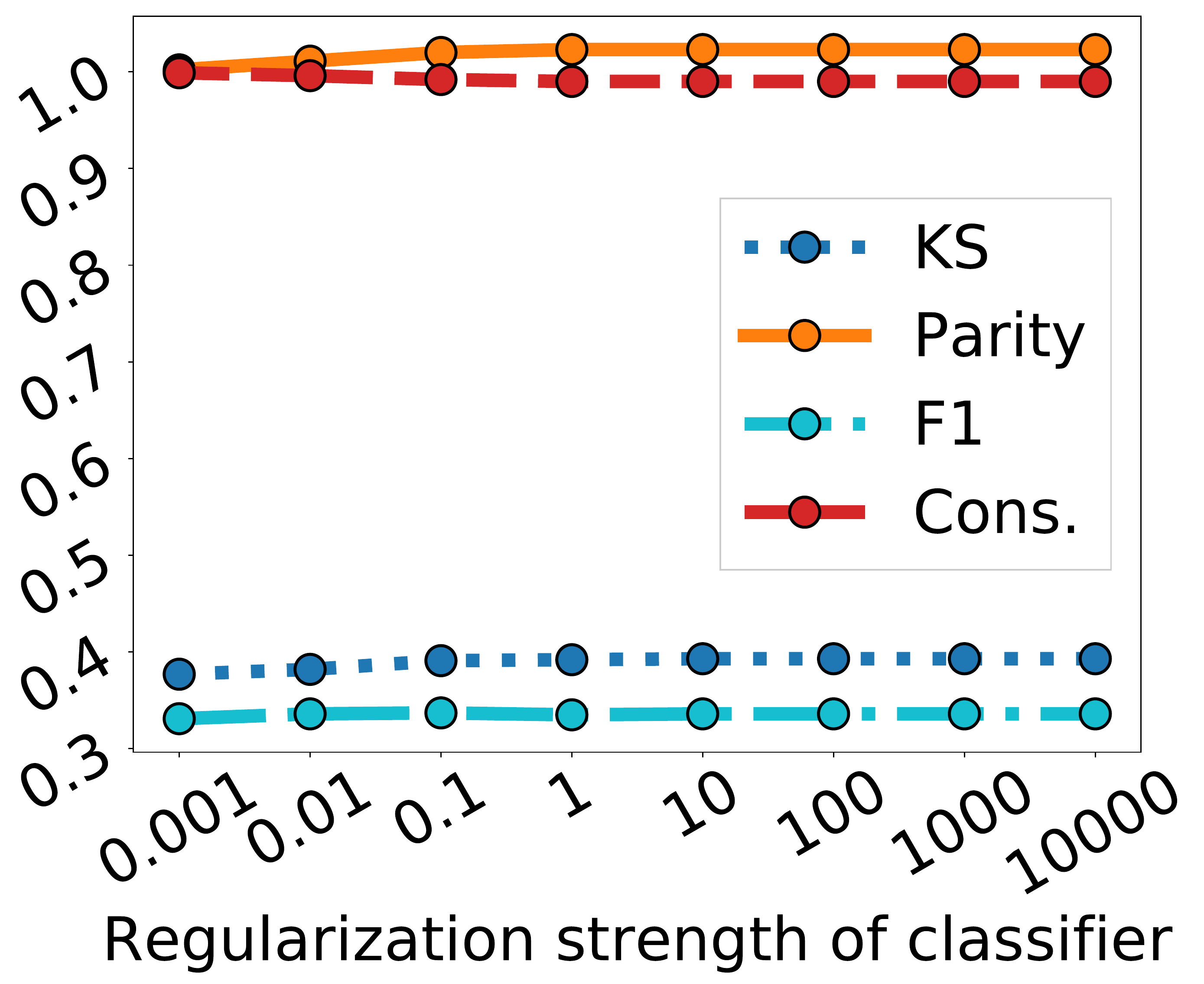}
	}
	\subfigure[Fraud]{
		\includegraphics[width=.22\linewidth]{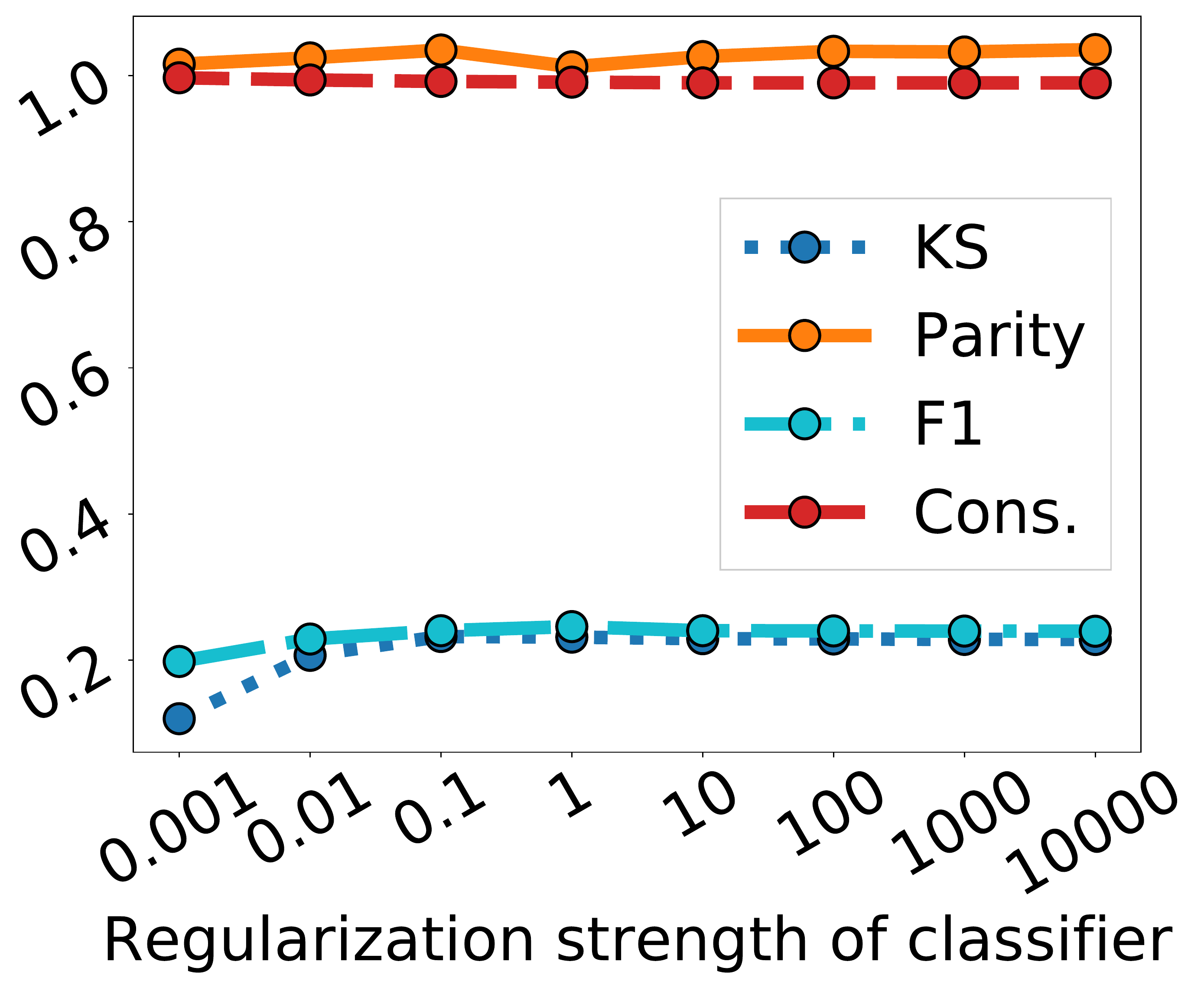}
	}
	\caption{Inverse of regularization strength of the classifier and the performance. 
		\small The x-axis is the inverse of the regularization strength; higher values means weaker regularization. The right-most value is proximal to the case where there is no regularization at all.
		\normalsize }
	\label{fig:params:cls}
\end{figure*}

We analyzed the influence of several hyperparameters, including the dimension of the feature space, the trade-off parameter \(\alpha\), and the regularization strength of the linear classifier, on information preservation, classification performance, and fairness. When studying one of them in particular, other hyperparameters are fixed, as described earlier in this section.

Figure~\ref{fig:params:dim} shows the influence of dimension. As we observed in experiments, a sufficiently high dimension, usually close to the original number of attributes, helps MSE and EMD reduction. In a higher dimension, MSE is more easily reduced, which makes it easier at reducing EMD as well. Therefore, we would suggest a sufficiently high dimension, often close to the dimension of the original attributes, unless there is need for dimensionality reduction. 

Figure~\ref{fig:params:alp} shows the trade-off between fairness and information preservation, controlled by \(\alpha\). Figure~\ref{fig:params:cls} shows the influence of the regularization strength of the classifier, which maps the samples in our learned feature space to the classification scores. Except for Adult, the influence of the regularization strength is mild, but the classification is still fairer when the regularization is stronger. Furthermore, the cost of achieving fairness in terms of classification results is not expensive.

\end{document}